\documentclass[10pt,twocolumn,letterpaper]{article}

\usepackage[pagenumbers]{cvpr} % To force page numbers, e.g. for an arXiv version

\definecolor{cvprblue}{rgb}{0.21,0.49,0.74}
\usepackage[pagebackref,breaklinks,colorlinks,allcolors=cvprblue]{hyperref}

\def\paperID{7600} % *** Enter the Paper ID here
\def\confName{CVPR}
\def\confYear{2026}

\title{AG-VAS: Anchor-Guided Zero-Shot Visual Anomaly Segmentation with Large Multimodal Models}

\author{
	Zhen Qu\textsuperscript{\rm 1, 2} \quad Xian Tao\textsuperscript{\rm 1,2,3,4}\footnotemark[2]~~ \quad Xiaoyi Bao\textsuperscript{\rm 1,2} \quad Dingrong Wang\textsuperscript{\rm 1, 2} \\ \quad ShiChen Qu\textsuperscript{\rm 1, 2} \quad Zhengtao Zhang\textsuperscript{\rm 1,2,3} \quad
	Xingang Wang\textsuperscript{\rm 1, 2} \\
	\textsuperscript{\rm 1}Institute of Automation, Chinese Academy of Sciences \\
	\textsuperscript{\rm 2}School of Artificial Intelligence, University of Chinese Academy of Sciences \\ \textsuperscript{\rm 3}Casivision \quad \textsuperscript{\rm 4} Weiqiao-UCAS Science and Technology Park
}

\usepackage{framed}
\usepackage{colortbl}
\usepackage{multirow}
\usepackage{wrapfig}
\usepackage{tabularx}
\usepackage{makecell}
\usepackage{enumitem}
\usepackage{marvosym}
\usepackage[accsupp]{axessibility}
\usepackage{float} % 加载 float 宏包
\definecolor{lightgreen}{RGB}{240,255,255}
\definecolor{darkred}{RGB}{180,0,0} % 定义暗红色

\definecolor{lightblue1}{RGB}{100, 149, 237} % Deeper blue
\definecolor{lightgreen1}{RGB}{34, 139, 34} % Deeper green
\definecolor{lightorange1}{RGB}{255, 165, 0} % Deeper orange

\definecolor{NOR}{RGB}{0, 176, 80} % Deeper blue
\definecolor{ANO}{RGB}{192, 0, 0} % Deeper green
\definecolor{SEG}{RGB}{112, 48, 160} % Deeper orange

\begin{document}
\maketitle
\footnotetext[2]{Corresponding Author: taoxian2013@ia.ac.cn.}
\footnotetext[3]{Project Page: \url{https://github.com/xiaozhen228/AG-VAS}}

\begin{abstract}
Large multimodal models (LMMs) exhibit strong task generalization capabilities, offering new opportunities for zero-shot visual anomaly segmentation (ZSAS). However, existing LMM-based segmentation approaches still face fundamental limitations: anomaly concepts are inherently abstract and context-dependent, lacking stable visual prototypes, and the weak alignment between high-level semantic embeddings and pixel-level spatial features hinders precise anomaly localization.
To address these challenges, we present AG-VAS (Anchor-Guided Visual Anomaly Segmentation), a new framework that expands the LMM vocabulary with three learnable semantic anchor tokens—[SEG], [NOR], and [ANO], establishing a unified anchor-guided segmentation paradigm. Specifically, [SEG] serves as an absolute semantic anchor that translates abstract anomaly semantics into explicit, spatially grounded visual entities (e.g., holes or scratches), while [NOR] and [ANO] act as relative anchors that model the contextual contrast between normal and abnormal patterns across categories. To further enhance cross-modal alignment, we introduce a Semantic-Pixel Alignment Module (SPAM) that aligns language-level semantic embeddings with high-resolution visual features, along with an Anchor-Guided Mask Decoder (AGMD) that performs anchor-conditioned mask prediction for precise anomaly localization.
In addition, we curate Anomaly-Instruct20K, a large-scale instruction dataset that organizes anomaly knowledge into structured descriptions of appearance, shape, and spatial attributes, facilitating effective learning and integration of the proposed semantic anchors. Extensive experiments on six industrial and medical benchmarks demonstrate that AG-VAS achieves consistent state-of-the-art performance in the zero-shot setting.
\end{abstract}    
\section{Introduction}
\label{sec:intro}

\begin{figure}[t]
	\centering
	\includegraphics[width= 1\columnwidth]{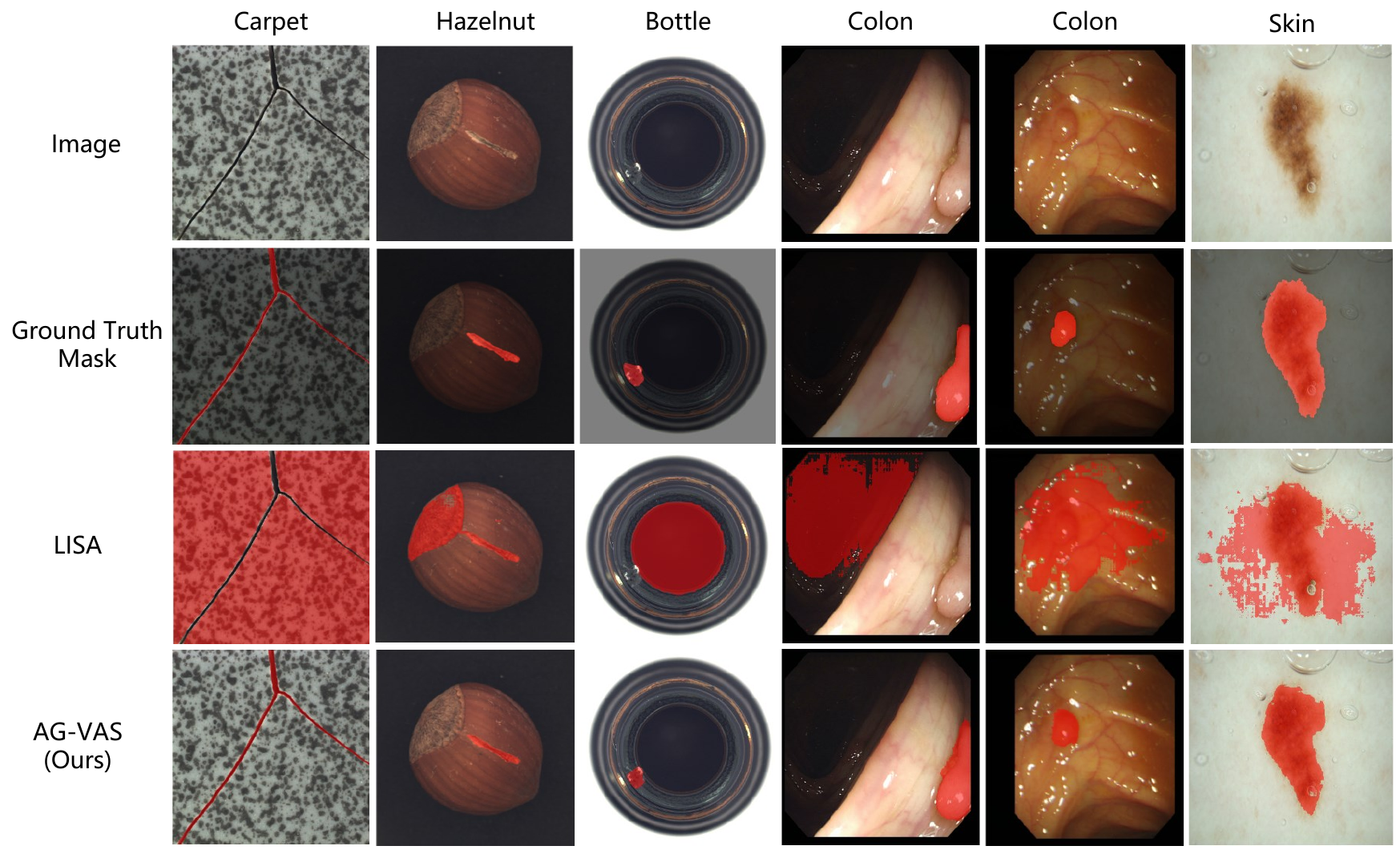}
	\caption{Comparison of zero-shot anomaly segmentation capabilities between LISA \cite{LISA} and our AG-VAS.}
	\label{Fig1}
\end{figure}

	\begin{figure*}[t]
	\centering
	\includegraphics[width= 2.0\columnwidth]{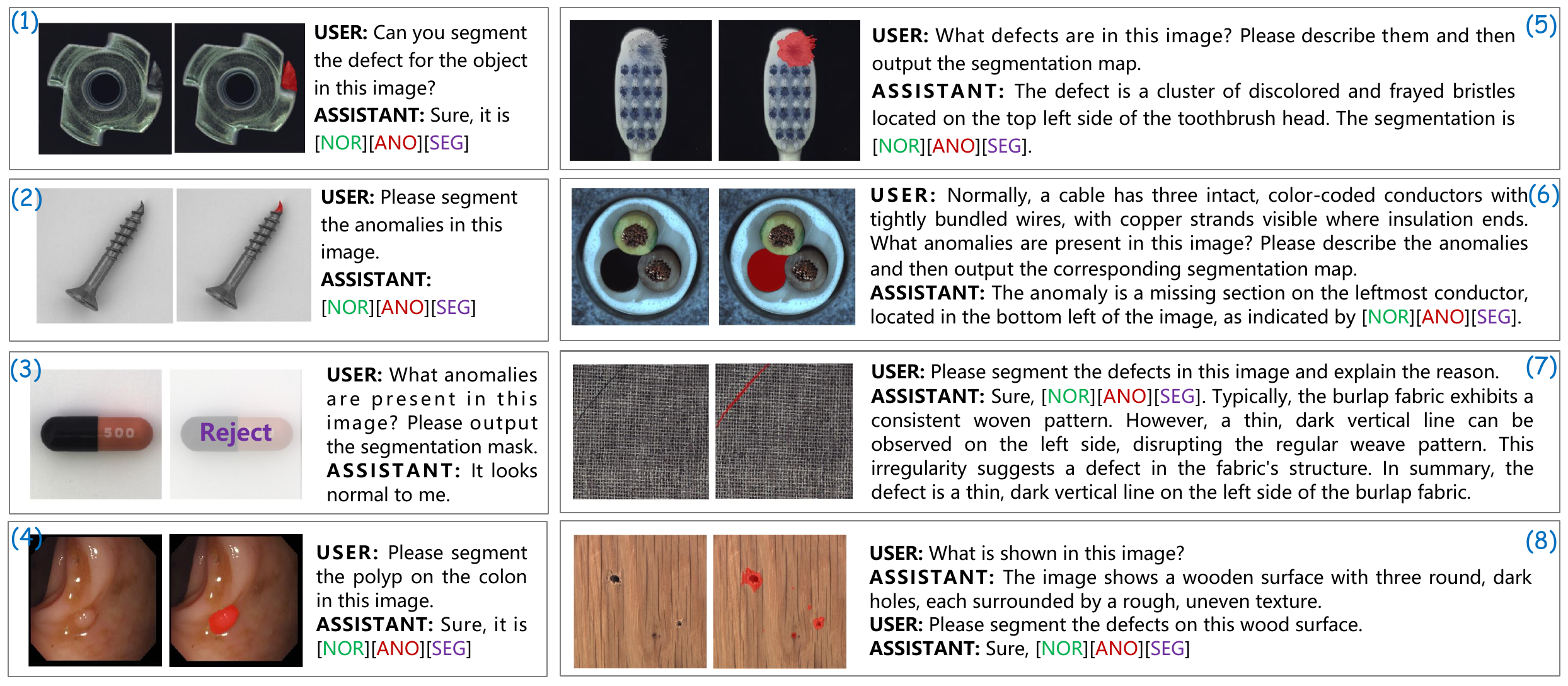}
	\caption{AG-VAS is an effective and efficient LMM for zero-shot anomaly segmentation and understanding. (Left: raw image; Right: segmentation result.) The visualization results are shown across the following scenarios: (1)–(4) direct segmentation, (5)–(6) describe-then-segment, (7) segment-then-explain, and (8) segmentation within dialogue.}
	\label{Fig2}
\end{figure*}

Zero-shot visual anomaly segmentation (ZSAS) aims to directly localize anomalous regions on the surfaces of unseen-category objects without the need for retraining. This capability is particularly important in data-scarce or privacy-sensitive scenarios, such as industrial defect localization \cite{industry1, CAPS, LSCAD} and medical image analysis \cite{medical1, Bayes, DictAS}. 
\par  
Most existing ZSAS models \cite{WinCLIP, AnomalyCLIP, AdaCLIP, Bayes, CLIP-AD, VCP, VAND} are built upon CLIP \cite{CLIP} due to its strong vision–language alignment capabilities. These models typically construct pairs of textual prompts representing normal and abnormal states, and align them with pixel-level patch embeddings to localize anomalies. Representative methods, such as AnomalyCLIP \cite{AnomalyCLIP} and Bayes-PFL \cite{Bayes}, further enhance the fine-grained anomaly perception via prompt learning strategies. However, constrained by the limited representation and understanding capabilities of CLIP, the ZSAS performance of these approaches appears to have reached a bottleneck. Moreover, they do not naturally produce binary segmentation masks, often requiring heuristic thresholding or empirical tuning, which hinders practical deployment. These limitations motivate exploring large multimodal models (LMMs) with stronger reasoning capacity and richer world knowledge \cite{LLaVA, InterVL, CogVLM, MiniGPT, PixelLM}.
\par 
Recent anomaly detection studies \cite{AnomalyGPT, AnomalyOV, EMIT, IADR1} leveraging LMMs have demonstrated remarkable performance in high-level visual understanding and conversational interaction. However, most existing works (e.g. Anomaly-OV \cite{AnomalyOV}) focus on enhancing image-level textual descriptions or anomaly classification accuracy, while precise pixel-level responses, such as binary segmentation masks, remain largely unexplored.  In this work, we aim to develop an end-to-end ZSAS framework built upon the pretrained LMMs, capable of understanding complex segmentation instructions and directly outputting binary masks.
\par
LISA \cite{LISA} is the first general-purpose image segmentation work to employ an \textit{embedding-as-mask} scheme, extending visual comprehension to pixel-wise segmentation by integrating LMMs with dedicated segmenters (e.g. SAM \cite{SAM}). However, when applied to industrial or medical ZSAS tasks, it often fails to correctly identify anomalous regions, sometimes even confusing foreground and background. For instance, given the instruction \texttt{"Please segment the anomalies in this image"}, LISA may produce masks that mislocalize the abnormal regions, as illustrated in Figure \ref{Fig1}. We attribute these failures in ZSAS to two main factors: 1) anomalies or defects are highly abstract concepts without concrete semantic references—for example, unlike a well-defined object such as ``apple'', anomalies can include ``holes'' or ``scratches'', making them difficult to map from textual descriptions; 2) the segmenter’s pixel-level features are not well aligned with the LMM’s visual-textual embedding space, limiting effective anomaly localization. 
\par
To address the above issues, we propose \textbf{AG-VAS}, an end-to-end ZSAS framework equipped with a set of learnable anchor tokens that encode contextual and world knowledge about anomalies. These anchors serve as semantic bridges between the LMM and the segmenter, enabling effective instruction-driven anomaly segmentation. Specifically, we extend the LMM vocabulary with special anchor tokens: the absolute semantic anchor [SEG] provides explicit cues regarding anomaly appearance, shape, and location, while the relative semantic anchors [NOR] and [ANO] capture contextual contrasts between normal and abnormal regions across diverse categories. To enhance cross-modal correspondence, we further introduce a \textit{Semantic–Pixel Alignment Module} (SPAM), which aligns high-level semantic embeddings from the LMM with fine-grained pixel features from the segmenter. Finally, an \textit{Anchor-Guided Mask Decoder} (AGMD) leverages these aligned representations for anchor-consistent mask generation, achieving binarized segmentation results.
\par 
\begin{figure*}[t]
	\centering
	\includegraphics[width= 1.8\columnwidth]{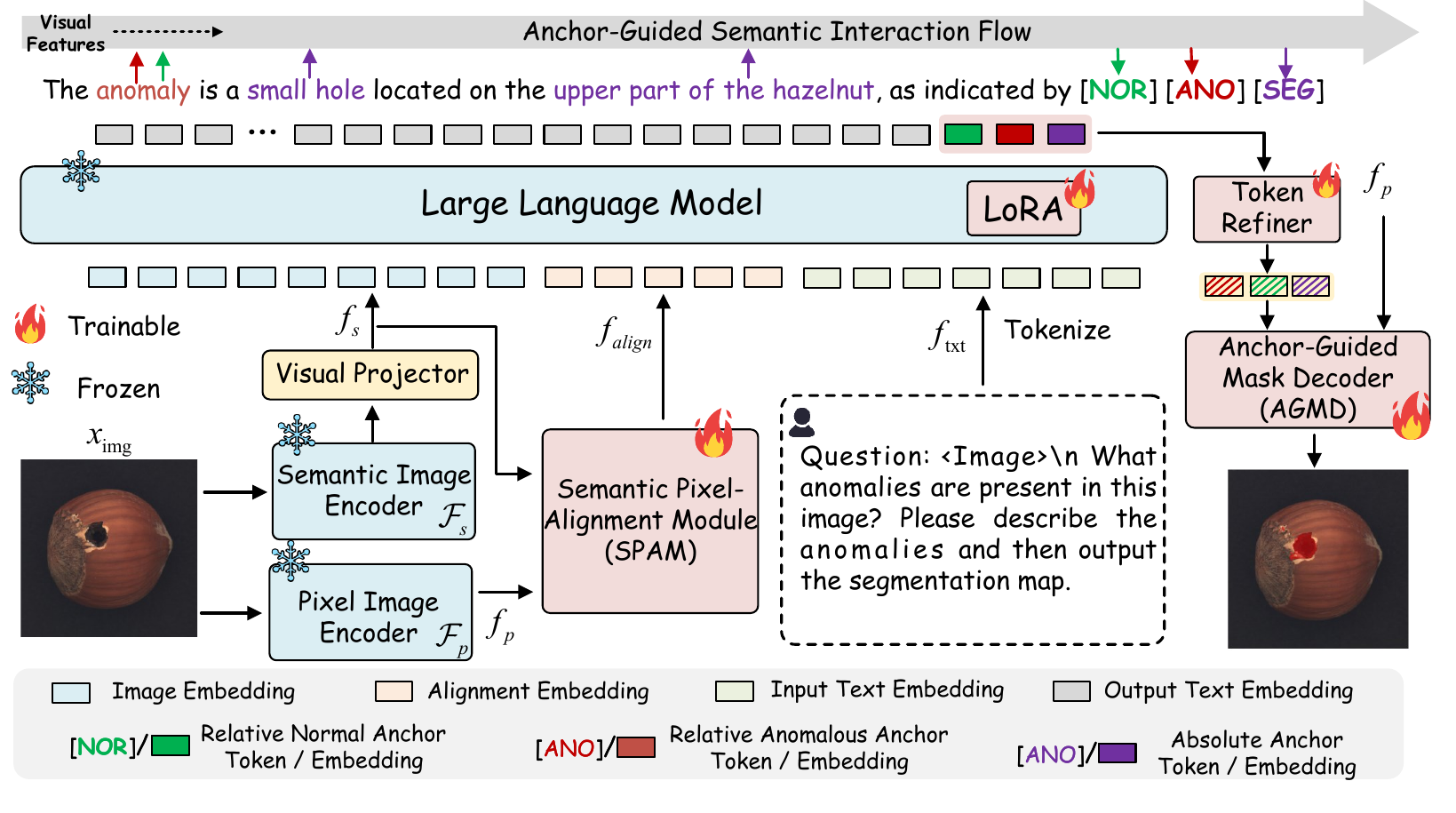}
	\caption{The pipeline of AG-VAS. We design two types of semantic anchors: the absolute anchor [\textcolor{SEG}{SEG}], which injects world and contextual knowledge about the appearance, structure, and location of anomalies (e.g., holes, scratches) into the segmentation process, and the relative normal/anomalous anchors [\textcolor{NOR}{NOR}]/[\textcolor{ANO}{ANO}], which guide the model to understand anomalies in relation to normal patterns. All parameters except those for the semantic/pixel image encoder, visual projector and LLM are trainable.}
	\label{Fig3}
\end{figure*}
To further enrich the model's understanding of anomaly-related semantics, we construct \textbf{Anomaly-Instruct20K}, an instruction-tuning dataset specifically designed for segmentation tasks. Unlike conventional instruction datasets focused on dialogue or visual question answering, Anomaly-Instruct20K injects structured world knowledge of anomalies into the LMM through diverse descriptions of defect appearance, shape, and spatial context. This enables the anchor tokens to learn precise and discriminative semantic representations. Moreover, its instruction-based design allows AG-VAS to develop interactive segmentation behaviors and flexibly respond to various user prompts. Figure~\ref{Fig2} demonstrates that AG-VAS produces high-quality masks under different segmentation instructions. In summary, our contributions are:
\par 
\begin{itemize}
	\item We propose \textbf{AG-VAS}, an anchor-guided framework for ZSAS that directly produces binary masks through instruction-driven segmentation. It introduces absolute and relative semantic anchors to bridge the LMM and segmenter, and incorporates two modules: SPAM for aligning LMM and segmenter features, and AGMD for generating accurate, anchor-consistent masks.
	\item We construct Anomaly-Instruct20K, a segmentation-oriented instruction-tuning dataset that injects structured world knowledge of anomalies into the LMM, enhancing its understanding of defect-related semantics.
	\item We achieve state-of-the-art performance on multiple industrial and medical anomaly detection benchmarks, demonstrating the strong generalization ability and practical applicability of AG-VAS.
\end{itemize}
%-------------------------------------------------------------------------
\section{Related Work}
\subsection{Large Multimodal Models}
Recent advances in LMMs can be broadly categorized into two paradigms: comprehension-oriented and perception-oriented approaches. Comprehension-oriented methods focus on enhancing visual understanding by aligning visual encoders with language models and improving representation quality through progressive training or multi-scale feature design \cite{BLIP2, Flamingo, llama_adapter, InstructBLIP, monkey, LLaVA, qwenVL}. Representative models such as BLIP-2 \cite{BLIP2}, Flamingo \cite{Flamingo}, LLaVA \cite{LLaVA}, and Qwen-VL \cite{qwenVL} have substantially advanced cross-modal reasoning and instruction-following capabilities. More recently, techniques like chain-of-thought reasoning \cite{COT1, COT2} and mixture-of-experts architectures \cite{MOE1, MOE2} have further improved LMMs’ ability to handle complex visual comprehension tasks. In contrast, perception-oriented LMMs emphasize fine-grained visual grounding, aiming to adapt language models for pixel-level localization tasks \cite{LISA, PixelLM, PaDT, LIRA, CORES, GSVA, omg_llava}. A promising direction is the embedding-as-mask paradigm, first introduced by LISA \cite{LISA} and later extended by PixelLM \cite{PixelLM} and PaDT \cite{PaDT}. While these methods achieve strong performance in general object segmentation, they remain suboptimal when applied to the ZSAS task.

\subsection{Zero-Shot Anomaly Segmentation}
Recent studies on ZSAS have highlighted the potential of foundation models such as CLIP \cite{CLIP} for localizing anomalies in unseen categories. Most existing methods \cite{WinCLIP, AdaCLIP, AnomalyCLIP, VAND, VCP, Bayes, AA, FAprompt} leverage CLIP's strong vision–language alignment and improve fine-grained anomaly understanding through feature adaptation or prompt learning. Early works like APRILN-GAN \cite{VAND} and CLIP-AD \cite{CLIP-AD} introduce learnable linear layers to project image patch features into the joint embedding space. Later methods strengthen cross-category ZSAS performance through various prompt-learning designs, including static learnable prompts (AnomalyCLIP \cite{AnomalyCLIP}), visual-context prompts (VCP-CLIP \cite{VCP}), hybrid static–dynamic prompts (AdaCLIP \cite{AdaCLIP}), and prompt distribution learning (Bayes-PFL \cite{Bayes}). Several LMM-based methods have recently been proposed for zero-shot tasks \cite{AnomalyOV, AnomalyGPT, IADR1, EMIT}, such as Anomaly-OV \cite{AnomalyOV} and IAD-R1 \cite{IADR1}. However, they primarily focus on image-level anomaly understanding and classification rather than segmentation. In contrast to the above methods, this work aims to develop an instruction-following ZSAS model based on LMMs that directly outputs binarized segmentation masks, which is more practical in real industrial and medical scenarios.

\section{Method}
Following the recent ZSAD setting \cite{AnomalyCLIP, Bayes}, our model is trained under supervision on seen categories $C^s$ from an auxiliary dataset including our curated Anomaly-Instruct20K, and directly tested on unseen categories $C^u$ collected from industrial and medical domains. Importantly, $C^s$ and $C^u$ are disjoint, and the datasets used for training and evaluation exhibit substantial domain gaps.

\subsection{Model Design}

\textbf{Framework overview.} As illustrated in Figure \ref{Fig3}, AG-VAS is an anchor-guided framework that integrates absolute and relative semantic anchors into an instruction-driven segmentation pipeline. It consists of four key components: 1) a frozen semantic image encoder $\mathcal{F}_s$ and pixel image encoder $\mathcal{F}_p$; 2) a large language model $\mathcal{F}_{\text{LLM}}$; 3) a Semantic–Pixel Alignment Module (SPAM); and 4) a lightweight Anchor-Guided Mask Decoder $\mathcal{D}$.
\par 
Given an input image $x_\text{img}$ and a text query $x_\text{txt}$, the frozen encoders first extract semantic features $f_s$ and pixel-level features $f_p$. These features are processed by the LLM together with the anchor tokens [NOR]/[ANO] and [SEG]. The mask decoder $\mathcal{D}$ then decodes the final-layer embedding corresponding to each anchor token to produce the binary segmentation mask $\mathbf{M}$. SPAM enhances this process by aligning the LLM's semantic representations with the pixel-level features, enabling the anchor embeddings to effectively guide mask generation. In the following, we describe each component in detail. 
\par 
\textbf{Design of Semantic Anchors.} In LMM-based ZSAS, we define \emph{semantic anchors} as bridges that connects the high-level semantic representations produced by the LMM with the mask decoder. The key motivation is to enable the model to capture fine-grained anomaly-related semantics through specific anchor embeddings in the LMM output space. From the perspective of a human inspector, locating anomalies in unseen categories generally relies on two complementary reasoning processes: 1) leveraging prior \emph{world knowledge} and contextual understanding of common defect types (e.g., holes, cracks, scratches) to recognize what anomalies might look like and where they typically appear; and
2) contrasting candidate regions against surrounding normal areas to identify inconsistencies or irregular patterns.
\par
Inspired by these two reasoning modes, we introduce two types of semantic anchors:
the \textbf{absolute anchor} [\textcolor{SEG}{SEG}], which encodes world knowledge and contextual cues regarding the appearance, structure, and location of potential anomalies, and the \textbf{relative anchors} [\textcolor{NOR}{NOR}]/[\textcolor{ANO}{ANO}], which guide the model to perceive anomalies in relation to normal patterns. These anchors are implemented as special tokens that are inserted into the text sequence and added in the LMM’s vocabulary space, allowing them to learn task-specific embeddings that differ from those obtained through conventional prompt tuning. During inference, AG-VAS mainly supports two segmentation modes that differ in whether the reasoning process is explicitly externalized:
\par 
(1) \emph{Implicit Mode}. The model directly segments anomalies by leveraging internalized contextual and world knowledge. In this work, this corresponds to the Direct Segmentation scenario. For example:
\vspace{-0.1in}
\begin{leftbar}
	\noindent
	{\small \textit{\textbf{User:} Please segment the anomalies in this image.}}
	\noindent
	
	\noindent{\small \textit{\textbf{Assistant:} Sure, it is [\textcolor{NOR}{NOR}][\textcolor{ANO}{ANO}][\textcolor{SEG}{SEG}].}}
\end{leftbar}
\vspace{-0.1in}
\par 
  (2) \emph{Explicit Mode}. The model externalizes its reasoning process either before or after mask prediction, such as describing anomalies before segmentation (Describe-then-Segment) or explaining the predicted mask (Segment-then-Explain). As an example, in the Describe-then-Segment process:
\vspace{-0.1in}
\begin{leftbar}
	\noindent
	{\small \textit{\textbf{User:} What anomalies are present in this image? Please describe the anomalies and then output the segmentation map.}}
	\noindent
	{\small \textit{\textbf{Assistant:} The anomaly is a small hole located on the upper part of the hazelnut, as indicated by [\textcolor{NOR}{NOR}][\textcolor{ANO}{ANO}][\textcolor{SEG}{SEG}].}}
\end{leftbar}
\vspace{-0.1in}
\noindent 
Both modes rely on identical anchor representations and world knowledge grounding; they differ only in whether the description or reasoning process is explicitly articulated. Unless otherwise specified, we adopt the Direct Segmentation mode for quantitative evaluation.
\par
\textbf{Semantic–Pixel Alignment Module.} The introduction of semantic anchors bridges the high-level semantic understanding of LMMs to pixel-level segmentation. However, a gap exists between the high-level semantic embedding space of the LMM and the low-level visual features of the pixel encoder, hindering the anchors' alignment with spatially coherent visual cues. Therefore, we propose the SPAM, which facilitates the alignment of pixel-level image features with the LMM’s embedding space through cross-modal interaction.
\par 
\begin{figure}[t]
	\centering
	\includegraphics[width= 1\columnwidth]{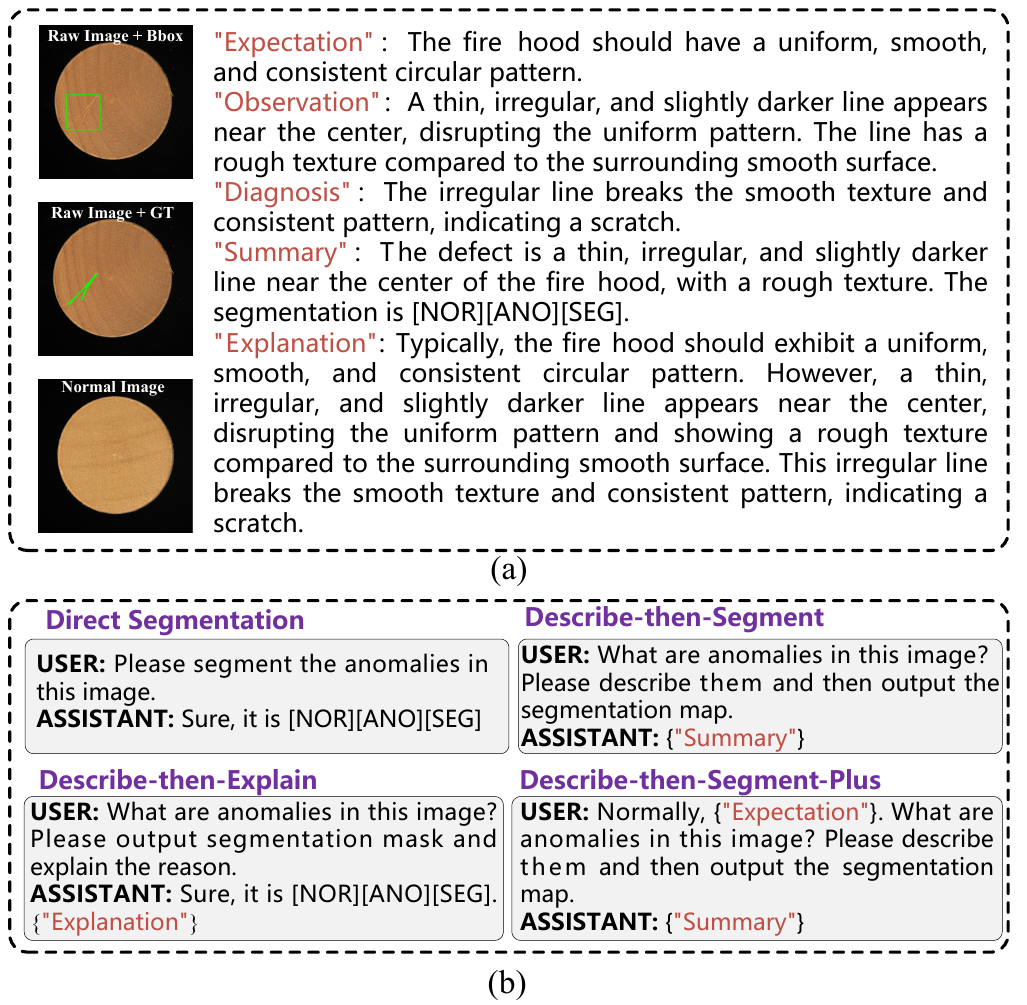}
	\caption{(a) An example from our Anomaly-Instruct20K dataset. (b) An example of instruction-tuning with dynamic fusion of templates during the actual training process. Note that only one specific template is shown here, with further details provided in the Appendix.
	}
	\label{Fig4}
\end{figure}
Specifically, given an image $x_\text{img} \in \mathbb{R}^{3\times h\times w}$, the semantic image encoder and the pixel image encoder are first used separately to extract visual features. The extracted features are then mapped to the same embedding dimension using a simple linear layer:
\begin{equation}
 f_s = Linear_s(\mathcal{F}_s(x_{\text{img}})) \quad f_p = Linear_p(\mathcal{F}_p(x_{\text{img}}))
\end{equation}
where $\mathcal{F}_s$ and $\mathcal{F}_p$ are semantic and pixel encoder, respectively. To facilitate better interaction between the anchors and the mask decoder, we employ a cross-modal attention module, where the semantic features extracted by the semantic image encoder serve as queries $\mathbf{Q}$ to attend to pixel embeddings ($\mathbf{K, V}$): 
\begin{equation}
	f_{align} = MHCA(\mathbf{Q} = f_s, \mathbf{K} = f_p, \mathbf{V} = f_p)
\end{equation}
where $MHCA$ denotes the multi-head cross attention \cite{MHCA}. Subsequently, the image embeddings $f_s$ from the semantic encoder, the aligned embeddings $f_{align}$, and the text embeddings $f_{\text{txt}}$ are concatenated and jointly fed into the large language model, which outputs a text response $\hat{y}_{\text{txt}}$:
\begin{equation}
	\hat{y}_{\text{txt}} = \mathcal{F}_{\text{LLM}}(Concat(f_s, f_{align},f_{\text{txt}} ))
\end{equation}
where $f_{\text{txt}}$ is a tokenized version of the input text query $x_{\text{txt}}$. The LLM output $\hat{y}_{\text{txt}}$ contains our designed anchor tokens [NOR],[ANO],[SEG].  We extract the LLM last-layer embeddings corresponding to these tokens and map them into the mask decoder space as $h_{nor}, h_{ano}, h_{seg}$ using a shared Token Refiner $\mathcal{F}_{tr}$, which consists of two linear layers.
\par 

\textbf{Anchor-Guided Mask Decoder.}
Given the refined anchor embeddings $h_{\text{nor}}, h_{\text{ano}}, h_{\text{seg}}$ and the pixel embeddings $f_p$, 
the proposed AGMD aims to generate absolute and relative segmentation masks guided by semantic anchors. We first concatenate the anchor embeddings with three learnable query tokens $\mathbf{T}_{\text{learn}} = \{t_{\text{nor}}, t_{\text{ano}}, t_{\text{seg}}\}$ to form the decoder input:
\begin{equation}
	\mathbf{Z}_0 = [t_{\text{nor}}, t_{\text{ano}}, t_{\text{seg}}, h_{\text{nor}}, h_{\text{ano}}, h_{\text{seg}}].
\end{equation}
The decoder performs $L$ layers of bidirectional cross-attention between the initial latent feature $\mathbf{Z}_0$ and pixel embeddings $f_p$, resulting in the final output $\mathbf{Z}_L$ and updated pixel embeddings $f_p'$. The process is described as:
\begin{equation}
	\mathbf{Z}_{L}, f_p' = \text{BiAttn}^L(\mathbf{Z}_0, f_p)
\end{equation}
where $\text{BiAttn}^L(\cdot)$ refers to applying a sequence of $L$ bidirectional cross-attention blocks, iteratively updating both $\mathbf{Z}_0$ and $f_p$, and following the design principles of SAM~\cite{SAM}.
After the final layer, we select the updated learnable tokens $\mathbf{T}'_{\text{learn}} = \{t'_{\text{nor}}, t'_{\text{ano}}, t'_{\text{seg}}\}$ from $\mathbf{Z}_{L}$ to compute segmentation masks. More architectural details of AGMD are provided in the Appendix.

For the absolute semantic anchor [SEG], a foreground probability map is produced via a sigmoid activation $\sigma(\cdot)$:
\begin{equation}
	P_{\text{seg}} = \sigma(t'_{\text{seg}} f_p'^{\top}),
\end{equation}
while for the relative anchors [NOR] and [ANO], a contrastive softmax function is applied to produce normal–abnormal probability maps:
\begin{equation}
	[P_{\text{nor}}, P_{\text{ano}}] = \text{Softmax}([t'_{\text{nor}}, t'_{\text{ano}}] f_p'^{\top}).
\end{equation}
The final anomaly map is obtained by fusing the absolute and relative maps:
\begin{equation}
	P = \alpha \cdot P_{\text{seg}} + (1-\alpha) \cdot P_{\text{ano}}, \quad \alpha=0.5
\end{equation}
Finally, the binary segmentation mask $\mathbf{M}$ is produced by thresholding the fused map at 0.5.
This design allows the decoder to transform abstract anchor semantics into pixel-level representations under the guidance of both absolute ([SEG]) and relative ([NOR]/[ANO]) anchors.
Detailed implementation and architecture specifications of AGMD are provided in the Appendix.

\subsection{Training Objectives}
The training of AG-VAS involves two complementary objectives:  
\par 
 1) \textbf{Textual Autoregressive Loss.}  
All tokens in the target sequence, including the anchor tokens, are supervised using the standard autoregressive cross-entropy loss:
\begin{equation}
	\mathcal{L}_{\text{txt}} = - \frac{1}{T_y}\sum_{t=1}^{T_y} \log P(y_t \mid y_{<t}, x_{\text{img}}, x_{\text{txt}})
\end{equation}
where $y_t$ denotes the ground-truth token at step $t$, $T_y$ is the length of the target sequence, and $x_{\text{img}}$, $x_{\text{text}}$ are the input image and text query, respectively.  
\par 
  2) \textbf{Segmentation Loss.}  
For the predicted masks from the Anchor-Guided Mask Decoder, we unify the loss for all anchors as a combination of Binary Cross-Entropy (BCE) loss and Dice loss:
\begin{equation}
	\mathcal{L}_{\text{seg}} = \sum_{c} \Big(\lambda_{bce} \text{BCE}(P_c, M_c) + \lambda_{dic}\text{Dice}(P_c, M_c) \Big)
\end{equation}
where $c\in\{\text{SEG}, \text{NOR}, \text{ANO}\}$,$P_c$ is the predicted mask for anchor $c$, and $M_c$ is the corresponding ground-truth mask. For the relative normal anchor, the ground-truth mask is the complement of the abnormal mask. The weights $\lambda_{bce}$ and  $\lambda_{dic}$ are set to 0.5 and 2.0 empirically.

The total training objective is then defined as:
\begin{equation}
	\mathcal{L} = \mathcal{L}_{\text{txt}} +  \mathcal{L}_{\text{seg}}
\end{equation}

\begin{table*}[t]
	\caption{Comparison with existing state-of-the-art methods. The best results are marked in \textcolor{darkred}{red}, while the second-best methods are indicated in \textcolor{blue}{blue}. The superscript * indicates that these methods were retrained on the anomaly segmentation dataset used in our AG-VAS.}
	\centering
	\label{Tab1}
	\renewcommand{\arraystretch}{1.1}
	\resizebox{2.0\columnwidth}{!}
	{
		\begin{tabular}{>{\centering\arraybackslash}p{1.6cm}>{\centering\arraybackslash}p{3cm}>{\centering\arraybackslash}p{3.2cm}*{6}{>{\centering\arraybackslash}p{2.5cm}}}  \toprule
			& \multirow{2}{*}{Method} & \multirow{2}{*}{Base Model} & \multicolumn{3}{c}{Industrial Datasets}                      & \multicolumn{3}{c}{Medical Datasets}                          \\    \cmidrule(lr){4-6}  \cmidrule(lr){7-9}
			&                         &                             & MVTec-AD           & KSDD2              & RSDD               & ISIC                & ColonDB            & ClinicDB           \\   \midrule
			\multirow{6}{*}{\makecell[c]{CLIP-Based \\ (AP, F1-Max, \\ $\text{IoU}_{\text{ano}}$)}} & WinCLIP \cite{WinCLIP}                 & CLIP(ViT-B-16-240)          & (18.0, 24.3, 0.0)  & (17.1, 24.6, 0.0)  & (2.1, 6.6, 0.1)    & (62.4, 64.1, 0.24)  & (14.3, 21.0, 0.0)  & (19.4, 27.3, 0.0)  \\
			& APRIL-GAN \cite{VAND}              & CLIP(ViT-L-14-336)          & (40.8, 43.3, 25.1) & (61.6, 62.6, 41.8) & (30.6, 40.2, 24.3) & (69.8, 69.3, 12.8)  & (23.2, 32.5, 14.4) & (38.8, 44.4, 25.3) \\
			& AnomalyCLIP \cite{AnomalyCLIP}            & CLIP(ViT-L-14-336)          & (34.5, 39.1, 22.6) & (41.8, 50.3, 27.6) & (19.1, 30.6, 8.8)  & (74.4, 70.8, 51.0)  & (31.3, 39.8, 25.2) & (42.2, 45.9, 30.0) \\
			& AnomalyCLIP* \cite{AnomalyCLIP}           & CLIP(ViT-L-14-336)          & (35.2, 39.0, 22.4) & (48.8, 53.6, 28.6) & (25.5, 34.4, 8.5)  & (76.5, 71.4, 54.1)  & (34.2, 39.8, 22.7) & (44.1, 47.3, 29.4) \\
			& Bayes-PFL \cite{Bayes}              & CLIP(ViT-L-14-336)          & (48.3, 49.1, 22.2) & (73.7, 63.5, 25.4) & (39.1, 45.2, 3.7)  & (84.6, 76.8, 60.0)  & (31.9, 39.2, 16.8) & (53.2, 51.7, 21.0) \\
			& Bayes-PFL* \cite{Bayes}              & CLIP(ViT-L-14-336)          & (\textcolor{blue}{50.3}, \textcolor{blue}{50.4}, 29.9) & (\textcolor{blue}{74.9}, 67.6, 39.5) & (41.9, 46.9, 7.7)  & (81.9, 74.3, 52.1)  & (30.5, 38.1, 27.7) & (47.6, 50.7, 34.4) \\  \midrule
			\multirow{9}{*}{\makecell[c]{LMM-Based \\ (AP, F1-Max, \\ $\text{IoU}_{\text{ano}}$)}}  & PixelLM \cite{PixelLM}                 & LLaVA-13B                   & (13.6, 19.6, 11.9) & (42.6, 46.8, 32.4) & (5.3, 9.4, 4.9)    & (69.8, 61.7, 48.9)  & (8.9, 16.8, 9.8)   & (13.0, 20.9, 14.6) \\
			& PaDT \cite{PaDT}                   & Qwen2.5-VL-7B               & (6.8, 12.9, 16.4)  & (1.2, 2.9, 12.4)   & (0.9, 1.8, 1.7)    & (65.9, 68.3, 54.5)  & (35.4, 48.4, 39.1) & (39.8, 51.4, 49.8) \\
			& LISA \cite{LISA}                   & LLaVA-7B                    & (9.8, 16.9, 10.2)  & (19.5, 23.8, 18.2) & (6.9, 12.5, 2.9)   & (68.1, 66.3, 54.6)  & (25.9, 33.1, 19.4) & (50.0, 45.3, 36.4) \\
			& LISA \cite{LISA}                   & LLaVA-13B                   & (14.6, 21.8, 13.7) & (36.8, 39.8, 29.4) & (5.2, 11.4, 2.6)   & (87.1, 78.1, \textcolor{blue}{64.8})  & (35.3, 34.0, 26.2) & (56.3, 52.9, 42.3) \\
			& LISA* \cite{LISA}                  & LLaVA-13B                   & (37.6, 40.8, 31.8) & (63.7, 63.8, 50.9) & (52.1, \textcolor{blue}{57.1}, 35.2) & (88.7, 82.4, 64.1)  & (40.9, 43.2, 29.5) & (67.8, 62.7, 48.3) \\
			& LISA \cite{LISA}                   & LLaVA-OneVision-7B          & (32.4, 38.3, 32.2) & (59.8, 65.8, 49.4) & (31.6, 37.6, 22.6) & (\textcolor{blue}{90.7}, 81.6, 56.9)  & (29.1, 35.8, 14.4) & (56.1, 54.4, 32.6) \\
			& LISA* \cite{LISA}                  & LLaVA-OneVision-7B          & (41.0, 44.1, 32.3) & (69.4, \textcolor{blue}{69.8}, 48.4) & (48.8, 54.9, 34.6) & (84.9, 77.6, 50.1) & (\textcolor{blue}{67.2}, \textcolor{blue}{62.9}, \textcolor{blue}{43.8}) & (\textcolor{blue}{81.0}, \textcolor{blue}{74.1}, \textcolor{blue}{59.8}) \\ \rowcolor{lightgreen}
			& AG-VAS                 & LLaVA-13B                   & (40.5, 42.5, \textcolor{blue}{37.0}) & (66.5, 65.4, \textcolor{blue}{53.7}) & (\textcolor{blue}{54.3}, 57.0, \textcolor{blue}{35.9}) & (88.8, \textcolor{blue}{82.7}, 64.6)  & (43.7, 45.1, 32.2) & (68.3, 65.5, 50.1) \\  \rowcolor{lightgreen}
			& AG-VAS                  & LLaVA-OneVision-7B          & (\textcolor{darkred}{51.0}, \textcolor{darkred}{52.7}, \textcolor{darkred}{44.8}) & (\textcolor{darkred}{76.6}, \textcolor{darkred}{74.7}, \textcolor{darkred}{53.8}) & (\textcolor{darkred}{56.2}, \textcolor{darkred}{57.8}, \textcolor{darkred}{36.9}) & (\textcolor{darkred}{92.7}, \textcolor{darkred}{84.9}, \textcolor{darkred}{65.0})  & (\textcolor{darkred}{70.7}, \textcolor{darkred}{66.2}, \textcolor{darkred}{58.2}) & (\textcolor{darkred}{86.6}, \textcolor{darkred}{79.2}, \textcolor{darkred}{69.5})   \\ \bottomrule
		\end{tabular}
	}
\end{table*}

\subsection{Anomaly-Instruct20K Dataset} 

To improve the instruction-based segmentation ability of large multimodal models, we introduce a new dataset, Anomaly-Instruct20K, designed to inject world knowledge on normal and abnormal visual patterns into the LMM backbone. Different from prior datasets such as Anomaly-Instruct-125K \cite{AnomalyOV}, which mainly provide question–answer annotations for general anomaly understanding, our dataset bridges vision–language reasoning with pixel-level anomaly perception through instruction-driven segmentation supervision. Anomaly-Instruct20K is built using samples from three publicly available datasets: RealIAD \cite{RealIAD}, GoodsAD \cite{Goods}, and DTD-Synthetic \cite{DTD}.
\par
We generate the dataset with the scientific multimodal model \textit{Inter-S1-241B}~\cite{intern}. 
For each anomalous sample, the model receives three visual inputs:
1) the raw image with bounding boxes, 
2) the raw image overlaid with the ground-truth mask, 
and 3) a corresponding normal reference image. 
Conditioned on these inputs, Inter-S1 generates a structured annotation comprising five fields: 
\textit{Expectation}, which describes the object's ideal normal appearance;  
\textit{Observation}, which details concrete visual deviations (e.g., location, shape, size, or texture irregularities);  
\textit{Diagnosis}, which explains why these deviations break normal material or structural consistency;  
\textit{Summary}, which converts the detected cues into a concise segmentation instruction and explicitly encodes spatial and appearance attributes necessary for grounding the anchor tokens [NOR][ANO][SEG];  
and \textit{Explanation}, which synthesizes all components into a coherent reasoning paragraph. 
An example of the generated annotations, along with the visual inputs, is shown in Figure~\ref{Fig4}(a).

During training, we generate final instructions by sampling templates from a library and combining them with dataset fields, covering four main instruction types: Direct Segmentation, Describe-then-Segment, Describe-then-Segment-Plus and Segment-then-Explain. An example of combining a template to generate a real instruction-tuning sample during the training process is shown in Figure 4(b). More detailed dataset construction procedures and system prompts are included in the Appendix.

\section{Experiment}
\subsection{Experimental Setup}
\textbf{Datasets.} The training of AG-VAS follows a multi-task joint learning strategy, leveraging datasets from two complementary domains: general object segmentation and anomaly segmentation. For general object segmentation, we adopt the setup from LISA \cite{LISA}, using datasets such as ADE20K \cite{App_Ade} to provide comprehensive segmentation capabilities. For anomaly segmentation, we employ two sources of auxiliary training data: 1) the constructed Anomaly-Instruct20K, and 2) a random set of 20k images extracted from existing industrial anomaly segmentation datasets \cite{GC, zju, Road, MIAD, DTD}, referred to as Anomaly-Seg20K, used under direct segmentation instructions. To preserve the model's dialogue and instruction-following capabilities, we include VQA datasets (LLaVA-150K \cite{LLaVA} and WebAD \cite{AnomalyOV}) during training. This joint training enables the absolute semantic anchor [SEG] to align with specific defect entities (e.g., holes, scratches) through contextual semantics and world knowledge, while the relative anchors [NOR] and [ANO] capture the semantic contrast between normal and anomalous regions. To evaluate the ZSAS performance of our model, we follow the protocol in \cite{Bayes} and conduct experiments on six real-world datasets, including three from the industrial domain (MVTec-AD \cite{MVTec}, KSDD2 \cite{KSDD2}, RSDD \cite{RSDD}) and three from the medical domain (ISIC \cite{ISIC}, ColonDB \cite{ColonDB}, ClinicDB \cite{ClinicDB}). More details about the datasets can be found in Appendix.
\par 
\textbf{Evaluation Metrics.} In this work, we focus on the quality of ZSAS predictions, particularly the binarized segmentation accuracy. We report pixel-level average precision (AP), the maximum F1 score at the optimal threshold (F1-Max), and the Intersection over Union of anomalous samples ($\text{IoU}_{\text{ano}}$). To further evaluate the model’s ability to reject normal images, we define $\text{IoU}_{\text{nor}}$: the model receives a score of 1 if it predicts an empty mask for a normal sample, and 0 otherwise. Both $\text{IoU}_{\text{ano}}$ and $\text{IoU}_{\text{nor}}$ are computed at the per-image level and then averaged over the corresponding subsets.
\par 
\textbf{Implementation Details.} By default, we adopt LLaVA-OneVision-7B \cite{llavaov} as our LMM backbone, equipped with a semantic image encoder (siglip-patch14-384), while SAM-ViT-H serves as the pixel image encoder for mask prediction. Training is implemented using DeepSpeed with the AdamW optimizer at a learning rate of 0.0003. During instruction tuning, the LLM is fine-tuned with LoRA at rank 16, using a global batch size of 80. The full training of the 7B model takes about 30 hours on four A100 GPUs with 40 GB memory each. More implementation details can be found in the Appendix.

\begin{table}[t]
	\caption{Comparison of the model's ability to reject segmentation on normal samples on the MVTec-AD.}
	\centering
	\label{Tab2}
	\renewcommand{\arraystretch}{1.2}
	\resizebox{1.0\columnwidth}{!}
	{
		\begin{tabular}{ccccc}
			\toprule
			Method  & Base Model         & $\text{IoU}_{\text{ano}}$ & $\text{IoU}_{\text{nor}}$ & Average \\ \midrule
			PixelLM \cite{PixelLM} & LLaVA-13B          & 11.9 & 2.4  & 7.2  \\
			PaDT \cite{PaDT}    & Qwen2.5-VL-7B      & 16.4 & 0.0  & 8.2  \\
			LISA \cite{LISA}    & LLaVA-7B           & 10.2 & 1.9  & 6.1  \\
			LISA \cite{LISA}    & LLaVA-13B          & 13.7 & 6.3  & 10.0 \\
			LISA* \cite{LISA}   & LLaVA-13B          & 31.8 & 43.5 & 37.7 \\
			LISA \cite{LISA}    & LLaVA-OneVision-7B & 32.2 & 4.0  & 18.1 \\
			LISA* \cite{LISA}   & LLaVA-OneVision-7B & 32.3 & 80.9 & 56.6 \\  \rowcolor{lightgreen}
			AG-VAS  & LLaVA-OneVision-7B & \textbf{44.8} & \textbf{87.7} & \textbf{66.3} \\  \bottomrule   
		\end{tabular}
	}
\end{table}

\subsection{Comparison with State-of-the-art methods}
Since LMM-based ZSAS is still in an early exploratory stage, directly comparable approaches are limited. Therefore, we include four representative CLIP-based ZSAS methods (WinCLIP \cite{WinCLIP}, APRIL-GAN \cite{VAND}, AnomalyCLIP \cite{AnomalyCLIP}, and Bayes-PFL \cite{Bayes}) together with three segmentation methods that address referring segmentation tasks, specifically designed for natural scene understanding: PixelLM \cite{PixelLM}, PaDT \cite{PaDT}, and LISA \cite{LISA}. To ensure a fair comparison, we also retrained other methods (e.g., AnomalyCLIP, Bayes-PFL, and LISA) on the anomaly segmentation dataset used by our AG-VAS during training. More details regarding the comparison of these methods are shown in Appendix.
\par  
\textbf{Quantitative Comparison.} Table \ref{Tab1} presents a comparison of our AG-VAS with other state-of-the-art methods across various industrial and medical datasets. It can be observed that when using the default base model, LLaVA-OneVision, our ZSAS performance surpasses all other methods on every metric. CLIP-based methods, such as Bayes-PFL, achieve comparable AP and F1-Max scores relative to LMM-based approaches but perform poorly on metric $\text{IoU}_{\text{ano}}$, which measures true segmentation accuracy. This suggests that their anomaly localization results are ambiguous, leading to lower $\text{IoU}_{\text{ano}}$ after binarization. Furthermore, we observe that LLM-based methods demonstrate superior zero-shot generalization on medical datasets compared to CLIP-based approaches, particularly after fine-tuning on our carefully curated Anomaly-Instruct20K and Anomaly-Seg20K datasets. It is noteworthy that no medical images were included in the training dataset, demonstrating that LMMs can leverage pre-existing world knowledge for zero-shot generalization to unseen domains. Furthermore, the introduction of the anchor mechanism allows our method to more effectively align semantic and pixel-level features, leading to substantial improvements over other LMMs in leveraging contextual semantics during training.
\par 
\textbf{Comparison of rejection performance.} For LMM-based ZSAS methods, it is crucial to not only accurately segment anomalous regions but also to reject normal samples by predicting empty masks, even when the user provides instructions to segment anomalies. The experimental results in Table \ref{Tab2} show that most LMMs struggle in this regard, whereas our AG-VAS demonstrates strong rejection ability, achieving an $\text{IoU}_{\text{nor}}$ of 87.7\% on normal samples. This indicates a substantial reduction in over-segmentation risk, while AG-VAS also maintains superior performance on anomalous regions, achieving an $\text{IoU}_{\text{ano}}$ of 44.8\%.

\subsection{Ablation}

\begin{table}[t]
	\caption{Ablation on different components.}
	\centering
	\label{Tab3}
	\renewcommand{\arraystretch}{1.2}
	\resizebox{1.0\columnwidth}{!}
	{
		\begin{tabular}{cccccc}
			\toprule
			&                          & AP   & F1-Max & $\text{IoU}_{\text{nor}}$ & $\text{IoU}_{\text{ano}}$ \\  \midrule
			\multirow{3}{*}{\makecell[c]{Module \\ Ablation}}             & w/o {[}SEG{]}            & 49.1 & 50.9   & 85.6 & 42.1 \\
			& w/o {[}NOR{]}{[}ANO{]}   & 47.2 & 49.7   & 52.1 & 39.5 \\
			& w/o SPAM                 & 46.5 & 48.0   & 70.6 & 41.4 \\  \midrule
			\multirow{3}{*}{\makecell[c]{Training  \\ Dataset \\ Ablation}} & w/o Anomaly-Instruct20K  & 48.4 & 50.0   & 85.2 & 39.9 \\
			& w/o Anomaly-Seg20K       & 49.3 & 51.5   & 53.5 & 42.4 \\
			& w/o General Segmentation & 36.1 & 36.5   & 70.2 & 34.9 \\    \rowcolor{lightgreen}\midrule
			& AG-VAS                   & \textbf{51.0} & \textbf{52.7}   & \textbf{87.7} & \textbf{44.8 } \\  \bottomrule  
		\end{tabular}
	}
\end{table}

In this subsection, ablation experiments on the MVTec-AD dataset are conducted to further investigate the impact of different settings on the proposed AG-VAS.

\textbf{Influence of Different Components.} Table~\ref{Tab3} presents an ablation study evaluating the impact of various modules and training datasets on ZSAS performance. From the module ablations, it can be observed that removing the absolute semantic anchor [SEG] leads to a noticeable drop in all metrics, highlighting its crucial role in aligning anchor semantics with concrete defect entities. Excluding the relative anchors [NOR] and [ANO] significantly decreases $\text{IoU}_{\text{nor}}$, indicating their importance in capturing the semantic contrast between normal and anomalous regions. The removal of SPAM also causes performance degradation, suggesting its effectiveness in feature aggregation. 

Regarding dataset ablations, training without Anomaly-Instruct20K leads to a noticeable decrease in performance, particularly on metrics reflecting anomaly segmentation. This indicates that instruction tuning with Anomaly-Instruct20K successfully injects domain-specific world knowledge into the model, enabling the anchors to better interpret and localize anomalies. Excluding Anomaly-Seg20K causes a larger drop in $\text{IoU}_{\text{nor}}$, showing that these auxiliary images are crucial for learning robust normal–abnormal contrasts. Omitting general segmentation data results in the most substantial decrease across all metrics, confirming that multi-domain joint training is essential for aligning absolute semantic anchors with visual features.
\par  

\textbf{Influence of different segmentation modes during inference.} Table~\ref{Tab4} compares AG-VAS performance under different inference strategies. \textit{Describe-then-Segment} refers to first generating a textual description of the target anomaly and then producing a segmentation mask. \textit{Describe-then-Segment-Plus} prepends an additional sentence describing the normal appearance of the object to the input instruction before the anomaly description. This serves as an expert-provided prior, explicitly informing the model about the object's typical state. \textit{Segment-then-Explain} first predicts the mask and then generates an explanation, while \textit{Direct Segmentation} predicts the mask directly from the instruction without intermediate text.

Compared with the default \textit{Direct Segmentation} mode, \textit{Describe-then-Segment} and \textit{Segment-then-Explain} slightly reduce segmentation performance due to textual noise introduced during intermediate reasoning. However, these modes enable the model to better follow user instructions and support interactive reasoning, justifying a minor sacrifice in mask accuracy. Notably, \textit{Describe-then-Segment-Plus}, by providing explicit knowledge of the object's normal state, improves contextual understanding and partially mitigates the performance drop, leading to the best overall results among the descriptive modes. 

\section{Conclusion}
This paper introduces AG-VAS, an anchor-guided framework for zero-shot visual anomaly segmentation that endows large multimodal models with learnable semantic anchors—absolute token [SEG] and relative tokens [NOR]/[ANO]—to jointly encode world knowledge about defect appearance, shape and location while simultaneously modeling the contextual opposition between normal and abnormal regions. By fusing high-level vision–language embeddings with fine-grained pixel features through the Semantic–Pixel Alignment Module, and decoding anchor-consistent masks via the bidirectional Anchor-Guided Mask Decoder, AG-VAS overcomes the lack of structured anomaly semantics and directly outputs accurate binary masks without requiring category-specific retraining. Extensive experiments on six challenging industrial and medical benchmarks demonstrate that AG-VAS establishes a new state-of-the-art, significantly outperforming both CLIP-based and recent LMM-based competitors. It shows strong rejection performance on normal images and demonstrates robust generalization across unseen domains, providing a practical solution for real-world zero-shot anomaly segmentation task.

\begin{table}[t]
	\caption{Ablation on different segmentation mode during inference.}
	\centering
	\label{Tab4}
	\renewcommand{\arraystretch}{1.2}
	\resizebox{1.0\columnwidth}{!}
	{
	\begin{tabular}{ccccc}
		\toprule
	 Segmentation Mode	& AP   & F1-Max & $\text{IoU}_{\text{nor}}$ & $\text{IoU}_{\text{ano}}$ \\  \midrule
		Describe-then-segment                 & 49.3 & 51.3   & 58.8 & 43.6 \\
		\makecell[c]{Describe-then-segment-plus} & \textbf{51.9} & \textbf{53.0}   & 90.2 & \textbf{45.1} \\
		Segment-then-explain                  & 48.6 & 51.4   & \textbf{91.1} & 43.4 \\  \rowcolor{lightgreen}\midrule
		Direct Segmentation                   & 51.0 & 52.7   & 87.7 & 44.8  \\  \bottomrule  
	\end{tabular}
}
\end{table}

\section*{\textbf{Acknowledgement}}
This work was supported by the Shandong Provincial Natural Science Foundation (No. ZR2025LGY002), the National Natural Science Foundation of China (No. 62373350), the Youth Innovation Promotion Association CAS (No. 2023145) and the Beijing Nova Program (No. 20240484687).
{
    \small
    \bibliographystyle{ieeenat_fullname}
    \bibliography{main}
}

\clearpage
\onecolumn
\appendix
\begin{center}
	\large \textbf{Appendix of AG-VAS: Anchor-Guided Zero-Shot Visual Anomaly Segmentation with Large Multimodal Models}
\end{center}
\par

This appendix includes the following five parts: 1) the construction of the Anomaly-Instruct20K dataset in Section \ref{secA}; 2) additional experimental details in Section \ref{secB}; 3) further ablation studies and analyses in Section \ref{secC}; 4) more detailed qualitative and quantitative results in Section \ref{secD}; 5) limitations of our method in Section \ref{secE}.

\section{Construction of Anomaly-Instruct20K}
\label{secA}
\begin{figure*}[h]
	\centering
	\includegraphics[width=1.0\columnwidth]{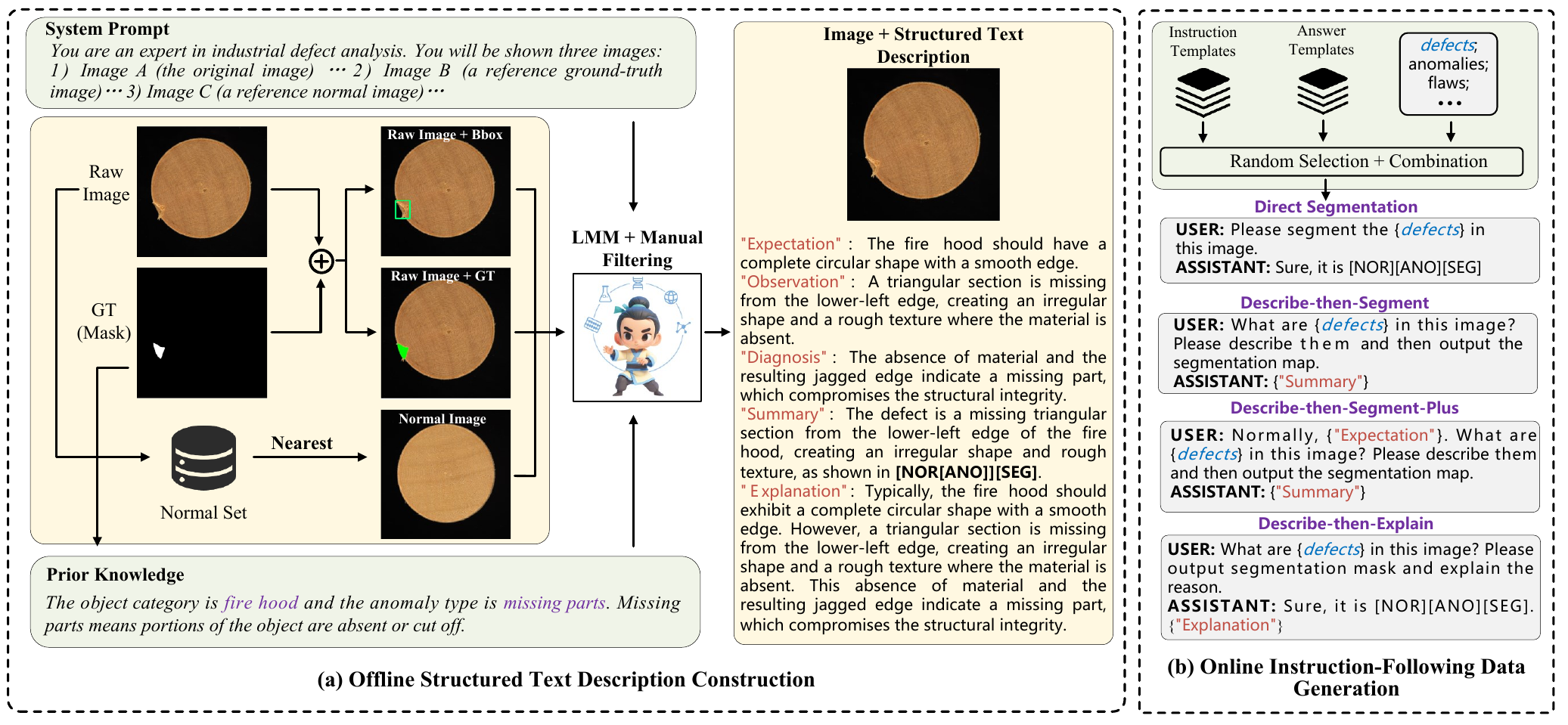}
	\caption{The Anomaly-Instruct20K generation pipeline for ZSAS. }
	\label{Fig_pipline}
\end{figure*}

\begin{figure*}[t]
	\centering
	\includegraphics[width=0.8\columnwidth]{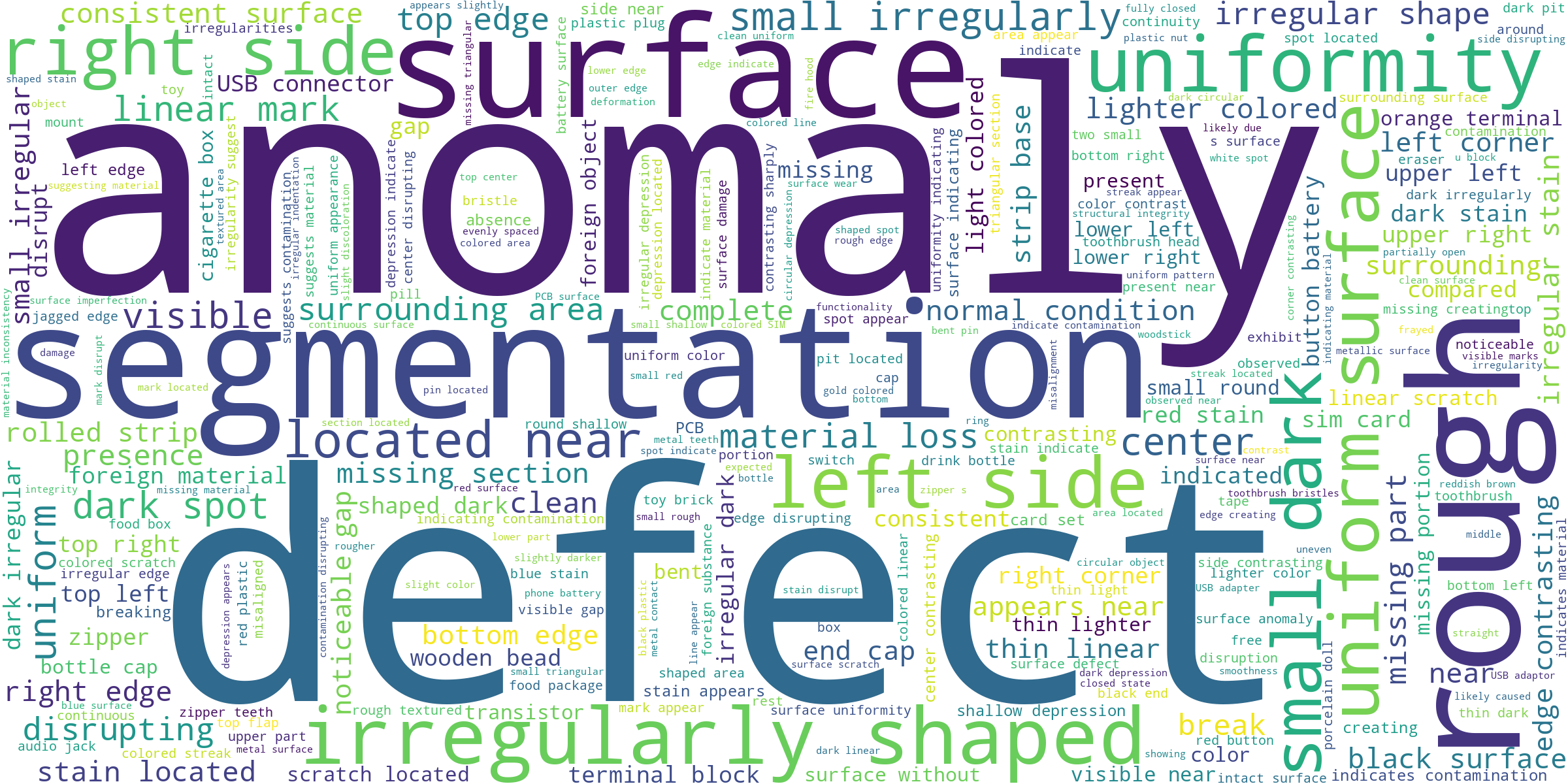}
	\caption{Word cloud of Anomaly-Instruct20K. Note that the special anchor words [NOR], [ANO], and [SEG] have been removed to reflect the true data distribution.}
	\label{Fig_word_cloud}
\end{figure*}
\subsection{Motivation of Anomaly-Instruct20K}
In this work, we construct Anomaly-Instruct20K, an instruction-tuning dataset specifically designed for zero-shot anomaly segmentation (ZSAS). The motivation for developing this dataset is twofold. First, it introduces rich world knowledge into existing large multimodal models (LMMs), enabling them to recognize anomalies, including their names, shapes, and locations, and to encode such information into the proposed anchor embeddings through semantic interaction flows within large language models (LLMs). Second, it equips the model with instruction-guided segmentation capability, allowing it to understand user intent, perform interactive reasoning, and ultimately produce accurate anomaly localization.
\par 
To construct Anomaly-Instruct20K, we use samples from three public datasets as data sources: RealIAD \cite{RealIAD}, GoodsAD \cite{Goods}, and DTD-Synthetic \cite{DTD}. Among these, RealIAD \cite{RealIAD} offers a large-scale industrial anomaly detection benchmark comprising 30 categories and multiple anomaly types, which serves as a key foundation for building our dataset. The GoodsAD \cite{Goods} dataset focuses on industrial goods, and the brand-dependent variations in product appearance contribute to a large expansion in the number of object categories. DTD-Synthetic \cite{DTD} is a specialized texture synthesis dataset that enhances the model’s ability to understand and handle texture-level anomalies.
\subsection{Pipeline for Constructing Anomaly-Instruct20K}
Table \ref{Fig_pipline} illustrates the construction pipeline of the proposed Anomaly-Instruct20K dataset, which consists of two stages: 1) offline generation of structured text descriptions from raw images using LMMs (e.g., fields such as ``Expectation'' and ``Observation'') to provide source data; and 2) online generation of instruction-following data during training, where carefully designed templates are applied to the source data to produce realistic fine-tuning samples for various segmentation tasks. 
\par 
\textbf{Offline Structured Text Description Construction.} Unlike existing approaches \cite{MMAD, PixelLM} that use LMMs to generate a single text description and then convert it into task-specific datasets, we aim to construct structured annotations for each image, as illustrated in Figure \ref*{Fig_pipline} (a). Specifically, the structured data is designed to cover the following five aspects: \textit{Expectation}, which describes the object's ideal normal appearance;  \textit{Observation}, detailing concrete visual deviations (e.g., location, shape, size, or texture irregularities);
\textit{Diagnosis}, explaining why these deviations violate normal material or structural consistency;
\textit{Summary}, converting the detected cues into a concise segmentation instruction and explicitly encoding the spatial and appearance attributes required to ground the anchor tokens [NOR], [ANO], and [SEG]; and \textit{Explanation}, synthesizing all components into a coherent reasoning paragraph.
\par 
Our method fully leverages the image understanding and text generation capabilities of the science-oriented LMM Intern-S1-241B \cite{intern}. We employ carefully designed language prompts together with human filtering to ensure the reliability of the generated content. Compared to commercial closed-source models such as GPT-4V \cite{GPT4V}, Intern-S1 is not only open-source but also demonstrates stronger capability in understanding scientific data, making it particularly suitable for our task. To further enhance anomaly understanding, we provide Intern-S1 with both visual and textual prompts. For visual inputs, three components are jointly fed into the LMM: the original image with the anomalous region highlighted by a green bounding box, the original image with the anomalous region masked, and a normal reference image from the same category. The normal reference image is selected following MMAD \cite{MMAD}, using the SSIM score \cite{SSIM} and Bhattacharyya distance \cite{SSIM2} to retrieve the most similar sample. For textual inputs, in addition to the system prompt, we provide prior knowledge in the form of a single sentence that specifies the object category, anomaly type, and corresponding explanation.
\par 
In designing the system prompt, we impose explicit constraints on the LMM’s textual outputs, such as prohibiting any direct reference to ground truth annotations or prior knowledge. In addition, we provide an in-context example to guide the model toward generating higher-quality and rule-compliant responses. Detailed formulations of the system prompt are presented in Figure \ref{system_prompts}. The final structured data further undergoes an additional round of manual verification, involving 10 annotators and requiring over 120 working hours.

Table \ref{Fig_word_cloud} illustrates the word cloud of Anomaly-Instruct20K. The results show that the dataset contains rich descriptions of anomaly categories (e.g., material loss), appearance attributes (e.g., dark spot), and spatial locations (e.g., left side). Such diverse semantic information facilitates LMMs in better understanding anomaly characteristics, thereby enabling more accurate localization.
\par
\textbf{Online Instruction-Following Data Generation.} After generating structured text descriptions for each image in the first step, how can we leverage these source data to train a model with instruction-following zero-shot anomaly segmentation (ZSAS) capability? As shown in Fig. \ref{Fig_pipline}(b), instruction-following data are generated online during training by combining the source data with randomly selected templates. For the three segmentation settings (direct segmentation, describe-then-segment, and segment-then-explain), we design dedicated sets of instruction templates together with their corresponding answer templates. To further enrich the variability of both instructions and responses, GPT-4 is employed to expand the template collections. 
\par 
Taking the describe-then-segment setting as an example, the instruction template \texttt{"What are the \{\textit{class\_name}\} in this image? Please describe them and then output the segmentation map"} is randomly selected from the instruction template set. The placeholder \{\textit{class\_name}\} is filled with a term randomly sampled from the anomaly target set $\mathcal{S}$, which includes entries such as \textit{defects}, \textit{anomalies}, and \textit{flaws}. For this task, the corresponding response is directly replaced with the \textit{Summary} from the source data, which inherently contains the anchor words [NOR],[ANO] and [SEG]. Subsequently, the image–instruction–response triplet is used as a training sample for instruction tuning. Detailed instruction and response templates can be found in Figures \ref{system_prompts},\ref{templates1},\ref{templates2}.

\begin{table*}[t]
	\caption{Construction of the Training Dataset.} 
	\centering
	\label{training_datasets}
	\renewcommand{\arraystretch}{0.9}
	\resizebox{0.9\columnwidth}{!}
	{
		\begin{tabular}{cccccc}
			\toprule
			Dataset                                          & Name          & Type                                        & Task                                                                                                       & Number                      & Sampling Rate         \\    \midrule
			\multirow{9}{*}{\makecell[c]{General   Segmentation \\ Datasets}} & ADE20K \cite{App_Ade}        & \multirow{2}{*}{\makecell[c]{Semantic \\ Segmentation}}      & Direct Segmentation                                                                                        & \multirow{9}{*}{$\sim$200k} & \multirow{9}{*}{0.4}  \\
			& COCO-Stuff \cite{App_cocostuff}    &                                             & Direct Segmentation                                                                                        &                             &                       \\   \cmidrule(lr){2-4}
			& PACO-LVIS \cite{App_Pacolvis}    & \multirow{3}{*}{\makecell[c]{Part Semantic \\ Segmentation}} & Direct Segmentation                                                                                        &                             &                       \\
			& PartImageNet \cite{App_PartImageNet} &                                             & Direct Segmentation                                                                                        &                             &                       \\
			& PASCAL-Part \cite{App_PascalPart}  &                                             & Direct Segmentation                                                                                        &                             &                       \\   \cmidrule(lr){2-4}
			& refCLEF \cite{App_refcocoplus}      & \multirow{4}{*}{\makecell[c]{Referring \\ Segmentation}}     & Direct Segmentation                                                                                        &                             &                       \\
			& refCOCO \cite{App_refcocoplus}      &                                             & Direct Segmentation                                                                                        &                             &                       \\
			& refCOCO+ \cite{App_refcocoplus}     &                                             & Direct Segmentation                                                                                        &                             &                       \\
			& refCOCOg \cite{App_refcocog}     &                                             & Direct Segmentation                                                                                        &                             &                       \\   \midrule
			\multirow{3}{*}{\textbf{Anomaly-Instruct20K}}             & RealIAD \cite{RealIAD}      & \multirow{3}{*}{\makecell[c]{Anomaly \\ Segmentation}}       & \multirow{3}{*}{\begin{tabular}[c]{@{}l@{}}Segment-then-explain\\      Describe-then-segment\end{tabular}} & \multirow{3}{*}{$\sim$20k}  & \multirow{3}{*}{0.25} \\
			& DTD-Synthetic \cite{DTD} &                                             &                                                                                                            &                             &                       \\
			& GoodsAD \cite{Goods}         &                                             &                                                                                                            &                             &                       \\   \midrule
			\multirow{6}{*}{Anomaly-Seg20K}                  & Road \cite{Road}         & \multirow{6}{*}{\makecell[c]{Anomaly \\ Segmentation}}       & \multirow{6}{*}{Direct Segmentation}                                                                       & \multirow{6}{*}{$\sim$20k}  & \multirow{6}{*}{0.25} \\
			& MIAD \cite{MIAD}         &                                             &                                                                                                            &                             &                       \\
			& GC \cite{GC}           &                                             &                                                                                                            &                             &                       \\
			& DTD-Synthetic \cite{DTD} &                                             &                                                                                                            &                             &                       \\
			& ZJU-Leaper \cite{zju}    &                                             &                                                                                                            &                             &                       \\
			& RealIAD \cite{RealIAD}     &                                             &                                                                                                            &                             &                       \\   \midrule
			\multirow{2}{*}{\makecell[c]{Visual Question \\ Answering Datasets}}                  & LLaVA-150k \cite{LLaVA}   & \multirow{2}{*}{\makecell[c]{Conversation}}  & General Conversation                                                                                       & \multirow{2}{*}{$\sim$222k} & \multirow{2}{*}{0.1}  \\
			& WebAD \cite{AnomalyOV}        &                                             & Anomaly Conversation                                                                                       &                             &                       \\ \bottomrule
		\end{tabular}
	}
\end{table*}

\section{Experimental Details}
\label{secB}
\subsection{Details of the Datasets}
\textbf{Training Datasets.} Table \ref{training_datasets} presents the auxiliary datasets used during the training of the proposed AG-VAS, which consist of four components: general segmentation datasets, Anomaly-Instruct20K, Anomaly-Seg20K, and visual question answering (VQA) datasets. For the general segmentation datasets, a setup similar to LISA \cite{LISA} is adopted, including semantic segmentation datasets \cite{App_Ade, App_cocostuff}, part semantic segmentation datasets \cite{App_Pacolvis, App_PartImageNet, App_PascalPart}, and referring segmentation datasets \cite{App_refcocog, App_refcocoplus}. They provide abundant samples for training the absolute semantic anchor [SEG], enabling it to more easily align with specific anomaly semantics, such as \textit{holes}, \textit{scratches}, and \textit{pits}. As described in Section A, Anomaly-Instruct20K provides the model with world knowledge, allowing it to comprehend what anomalies are, where they appear, and their visual characteristics. Additionally, it equips the LMM to perform both Segment-then-Explain and Describe-then-Segment tasks. For Anomaly-Seg20K, we randomly sampled 20k images from several industrial anomaly segmentation datasets \cite{Road, MIAD, GC, DTD, zju, RealIAD}, ensuring a balanced distribution of normal and anomalous samples. These datasets are used for direct segmentation tasks without requiring any additional annotations. Anomaly-Seg20K and Anomaly-Instruct20K are employed together to train the relative semantic anchors [NOR]/[ANO] and the absolute semantic anchor [SEG]. \textit{This allows the model to utilize image background information to detect relative anomalies, while simultaneously exploiting world knowledge and contextual semantics to precisely localize absolute anomalous regions or objects (e.g., holes, bent wires).} Finally, the VQA datasets are further utilized to preserve the model’s language description and dialogue capabilities, with LLaVA-150k \cite{LLaVA} used for general conversation and WebAD \cite{AnomalyOV} for anomaly-specific conversation.
\par 
\textbf{Testing Datasets.} Following \cite{Bayes}, we evaluate AG-VAS on six datasets spanning industrial and medical domains under the zero-shot anomaly segmentation (ZSAS) setting. In the industrial domain, MVTec-AD \cite{MVTec}, KSDD2 \cite{KSDD2}, and RSDD \cite{RSDD} are used for comprehensive evaluation. MVTec-AD, a widely adopted benchmark, contains 15 categories of textures and objects. KSDD2 and RSDD are real-world industrial datasets, where KSDD2 evaluates anomaly localization under low-contrast backgrounds, and RSDD focuses on defect detection on railway surfaces. In the medical domain, ISIC \cite{ISIC} is used for skin lesion analysis, while ClinicDB \cite{ClinicDB} and ColonDB \cite{ColonDB} are used for colon polyp segmentation. Notably, under the zero-shot setting, the training and testing categories are disjoint, with substantial domain gaps.

\subsection{Details of Implementations}
\textbf{Training details.} By default, we adopt LLaVA-OneVision-7B \cite{llavaov} as our LMM backbone, equipped with a semantic image encoder (siglip-patch14-384), while SAM-ViT-H \cite{SAM} serves as the pixel image encoder for mask prediction. Training is implemented using DeepSpeed \cite{deepspeed} with the AdamW optimizer \cite{AdamW} at a learning rate of 0.0003. The learning rate is scheduled using WarmupDecayLR, with the warmup period configured to 100 iterations. The weights $\lambda_{bce}$ and $\lambda_{dic}$ for the segmentation losses are empirically set to 0.5 and 2.0, respectively. During instruction tuning, the LLM is fine-tuned using LoRA \cite{lora} with rank 16 and a global batch size of 80. The trainable parameters include the LoRA weights, the proposed SPAM and AGMD modules, as well as the LLM token embeddings and output head. To address the imbalance among the four training data sources, we assign sampling probabilities of 0.4, 0.25, 0.25, and 0.1 at each iteration, as summarized in Table \ref{training_datasets}.  Due to the imbalance in the sizes of the four main training datasets, we assign sampling probabilities of 0.4, 0.25, 0.25, and 0.1 for each step to balance the different data types and training tasks, as summarized in Table \ref{training_datasets}. In addition, to enable the relative semantic anchors [NOR]/[ANO] to learn how to localize anomaly regions based on the surrounding image context, we train them exclusively on Anomaly-Instruct20K and Anomaly-Seg20K. The absolute semantic anchors, in contrast, are trained jointly using all segmentation datasets. The full training of the 7B model requires approximately 30 hours on four A100 GPUs with 40 GB memory each. 
\par 
\textbf{Testing details.} During inference, we by default employ the direct-segmentation approach, corresponding to the \textit{Implicit Mode} described in Section 3.1 of the main text. In this mode, the model uses its learned world knowledge and contextual semantics to perform implicit internal reasoning and directly output the segmentation mask of the target anomalous regions. Specifically, the default instruction used during inference is \texttt{"Please segment the anomalies in this image"}. After the LLM autoregressively generates a response containing the anchor tokens [NOR], [ANO], and [SEG], we extract the final-layer hidden states corresponding to these tokens to obtain the anchor embeddings. These embeddings are then fed into the proposed AGMD module to produce the segmentation probability maps $P_{\text{ano}}$ and $P_{\text{seg}}$. The two probability maps are then averaged using a weighting coefficient of $\alpha = 0.5$ to obtain the final anomaly map. Finally, a threshold of 0.5 is applied to obtain the binary segmentation map $\mathbf{M}$. 

\subsection{Details of Evaluation Metrics}
In this work, four metrics are employed to evaluate the model's ZSAS performance on novel categories. This includes two threshold-independent metrics, which have been widely adopted in previous CLIP-based methods \cite{AnomalyCLIP, Bayes, AdaCLIP}: average precision (AP) and the maximum F1 score at the optimal threshold (F1-Max). However, in practical industrial and medical anomaly segmentation scenarios, the objective is not merely to produce anomaly score maps, but to obtain accurate binary masks of anomalous regions. To this end, we further introduce the pixel-level intersection over union ($\text{IoU}_{\text{ano}}$) to quantify the overlap between the predicted binary mask $\mathbf{M}$ and the ground-truth mask $\mathbf{G}$ for anomalous samples:
\begin{equation}
	\text{IoU}_{\text{ano}} 
	= \frac{1}{N_\text{ano}} \sum_{i=1}^{N_\text{ano}} 
	\frac{|\mathbf{M}_i \cap \mathbf{G}_i|}{|\mathbf{M}_i \cup \mathbf{G}_i|},
\end{equation}
where $N_\text{ano}$ denotes the total number of anomalous samples in the test set, and $\mathbf{M}_i, \mathbf{G}_i \in \{0,1\}^{H\times W}$ are the binarized prediction mask and ground-truth mask for the $i$-th anomalous image, with $H$ and $W$ being the height and width of the image, respectively. 
\par 
To assess the model's ability to reject normal samples by predicting an empty mask, we introduce an additional metric, $\text{IoU}_{\text{nor}}$, which equals 1 if the predicted mask for a normal sample is empty, and 0 otherwise:
\begin{equation}
	\text{IoU}_{\text{nor}} 
	= \frac{1}{N_\text{nor}} \sum_{i=1}^{N_\text{nor}} 
	\mathbf{1}\{\mathbf{M}_i = \mathbf{0}\},
\end{equation}
where $N_\text{nor}$ denotes the number of normal samples in the test set, 
$\mathbf{M}_i$ is the predicted mask for the $i$-th normal image, 
$\mathbf{0}$ represents an empty mask (all pixels are zero), 
and $\mathbf{1}\{\cdot\}$ is the indicator function that returns 1 if the condition is true and 0 otherwise.
\subsection{Details of the Model Architecture}
\begin{wrapfigure}{r}{0.5\textwidth} 
	\centering
	%\setlength{\intextsep}{0pt}
	%\vspace{-25pt}
	%\hspace{-25pt}
	\includegraphics[width=0.5\columnwidth]{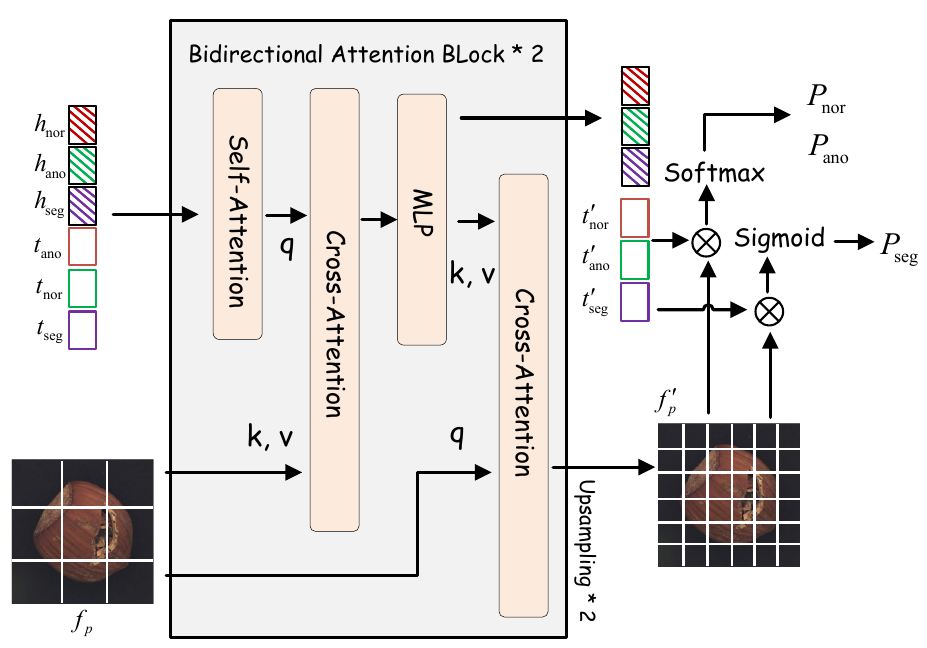} 
	\caption{The proposed AGMD architecture.}
	\label{AGMD}
\end{wrapfigure}
The proposed AG-VAS adopts LLaVA-OneVision-7B \cite{llavaov} as its default base LMM, with Qwen2 \cite{qwen2} as an alternative LLM. Since Qwen2's text embedding space has a default dimensionality of 3584, we use a Token Refiner composed of two linear layers to map the extracted anchor embeddings into the 256-dimensional SAM \cite{SAM} image embedding space. The refined anchor embeddings $h_\text{seg}, h_\text{nor}, h_\text{ano}$ are concatenated with three learnable vectors $t_\text{seg}, t_\text{nor}, t_\text{ano}$ and then fed into the proposed AGMD, enabling the segmentation targets to be located in the pixel embedding space based on the anchors. Figure \ref{AGMD} illustrates the architecture of the proposed AGMD. It primarily relies on two bidirectional attention blocks adapted from SAM \cite{SAM} to facilitate interaction between the anchor and pixel embeddings. This design allows the anchor embeddings from the LMM to more effectively absorb information from the pixel embedding space. Conversely, the pixel embeddings become better aligned with the semantic embedding space through cross-modal attention. Finally, the three output learnable vectors $t'_{\text{seg}}, t'_{\text{nor}}, t'_{\text{ano}}$, which have fully integrated information from both the anchor and pixel embeddings, are multiplied with the refined pixel features $f'_p$ and then passed through softmax and sigmoid activation functions to obtain the relative probability map $[P_{\text{nor}}, P_{\text{ano}}]$ and the absolute probability map $P_{\text{seg}}$.
\begin{table}[tb]
	\centering
	\begin{minipage}[t]{0.45\textwidth}
		\centering
		\caption{Ablation on different pixel image encoders on the MVTec-AD.}
		\label{pixel_encoder}
		\renewcommand{\arraystretch}{1.3}
		\resizebox{1\columnwidth}{!}
		{
			\begin{tabular}{ccccc}
				\toprule
				Pixel   Image Encoder & AP   & F1-Max & $\text{IoU}_{\text{nor}}$ & $\text{IoU}_{\text{ano}}$ \\ \midrule
				CLIP(ViT-L-14-336)    & 48.7 & 51.8   & 75.2     & 39.4     \\
				SigLIP(ViT-L-14-384)  & 46.8 & 48.5   & 72.1     & 37.1     \\
				DINOv2 (ViT-L-14)     & 53.2 & 53.8   & 56.5     & 44.1     \\
				DINOv2 (ViT-G-14)     & 57.8 & \textbf{57.3}   & 64.2     & 49.2    \\
				DINOv3 (ViT-H-16)     &\textbf{ 59.6} & 58.9   & 74.6     & \textbf{49.4}     \\
				SAM-FT (ViT-H-16)     & 39.9 & 40.6   & 60.4     & 32.3     \\   \rowcolor{lightgreen}\midrule
				SAM (ViT-H-16)        & 51.0 & 52.7   & \textbf{87.7}     & 44.8     \\   \bottomrule    
			\end{tabular}
		}
	\end{minipage}\hfill
	\begin{minipage}[t]{0.5\textwidth}
		\centering
		\caption{Comparison of average inference time and maximum GPU cost on MVTec-AD. The best results are shown in \textbf{bold}.}
		\label{efficiency}
		\renewcommand{\arraystretch}{1.7}
		\resizebox{1\columnwidth}{!}
		{
			\begin{tabular}{cccccc}
				\toprule
				Method  & Base Model         & AP   & $\text{IoU}_{\text{ano}}$ & GPU (GB) & Time (s) \\   \midrule
				PaDT    & Qwen2.5-VL-7B      & 6.8  & 16.4   & 19.4    & 1.2      \\
				PixelLM & LLaVA-13B          & 13.6 & 11.9   & 52.8    & 6.9      \\
				LISA    & LLaVA-7B           & 9.8  & 10.2   & \textbf{15.6}    & \textbf{0.6}      \\
				LISA*   & LLaVA-13B          & 37.6 & 31.8   & 27.8    & 1.5      \\
				LISA*   & LLaVA-OneVision-7B & 41.0 & 32.3   & 19.9    & 0.97     \\  \rowcolor{lightgreen}\midrule
				AG-VAS  & LLaVA-OneVision-7B & \textbf{51.0} & \textbf{44.8}   & 20.8    & 1.0       \\   \bottomrule    
			\end{tabular}
		}
	\end{minipage}
\end{table}
\section{Additional Ablations and Analysis}
\label{secC}

\subsection{Ablation on the Pixel Image Encoder}
Table \ref{pixel_encoder} reports the performance obtained by replacing the default SAM (ViT-H) \cite{SAM} with alternative pixel-level encoders. Here, SAM-FT denotes the variant where a linear layer is appended to the encoder output and fine-tuned during training. Results show that, in addition to SAM, other encoders such as CLIP \cite{CLIP}, DINOv2 \cite{dinov2}, and DINOv3 \cite{dinov3} also achieve strong performance, with DINOv2 and DINOv3 even outperforming SAM in terms of ZSAS. This suggests that our method is not tied to a specific pixel encoder and can flexibly adapt to different backbone choices. Moreover, as the representational capacity of the encoder increases, the zero-shot performance of AG-VAS consistently improves, highlighting the scalability of our framework. Notably, fine-tuning SAM with an additional linear layer leads to a significant performance drop, indicating that preserving the original pre-trained representations is crucial for maintaining generalization.

\subsection{Analysis of Inference Efficiency}
Table \ref{efficiency} compares different LMM-based methods in terms of ZSAS performance, maximum GPU memory consumption per image, and average inference time on the MVTec-AD dataset. Our method achieves the best AP and $\text{IoU}_{\text{ano}}$ while maintaining competitive inference efficiency. In particular, AG-VAS significantly outperforms prior LMM-based approaches such as LISA and PixelLM, demonstrating the effectiveness of the proposed modules. Despite the improved segmentation accuracy, AG-VAS introduces only a modest increase in GPU memory usage and inference time compared to the LISA baseline, indicating an efficient design. Overall, these results show that our method not only improves ZSAS performance but also remains practical for real-world deployment.

\subsection{Analysis of Failure Cases}
\begin{wrapfigure}{r}{0.55\textwidth} 
	\centering
	%\setlength{\intextsep}{0pt}
	%\vspace{-25pt}
	%\hspace{-25pt}
	\includegraphics[width=0.55\columnwidth]{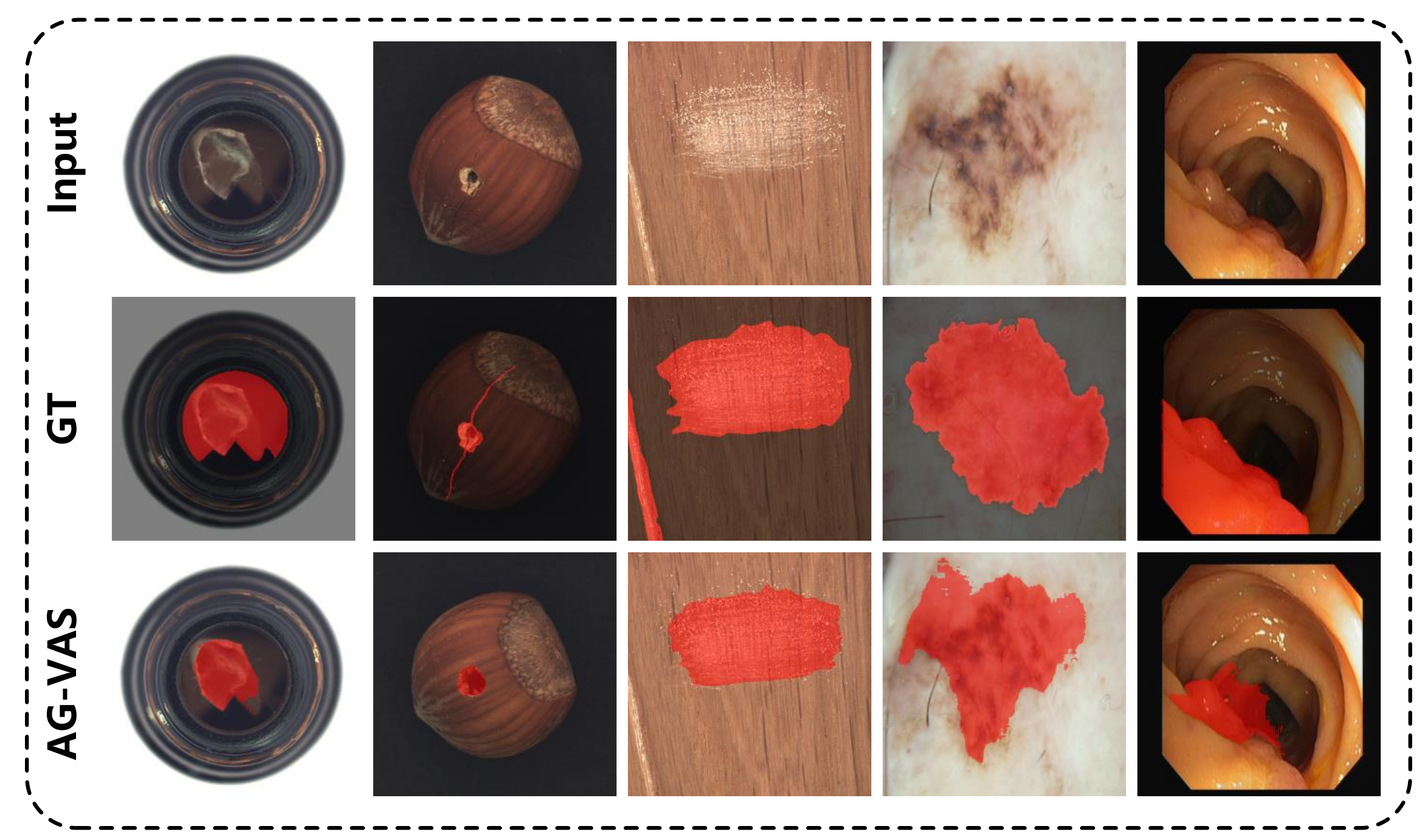} 
	\caption{Failure Cases.}
	\label{FC}
\end{wrapfigure}
Figure \ref{FC} presents failure cases of the proposed AG-VAS in the ZSAS task. It can be observed that AG-VAS tends to produce more conservative predictions; that is, for anomalous regions without clear boundaries, the binarized segmentation masks are smaller than the ground-truth regions. In addition, we find that AG-VAS performs poorly when segmenting multiple anomalous regions within a single image, especially when the anomaly types differ significantly. For example, in the second column (hazelnut), the crack outside the hole is ignored by the model, and in the third column (wood), the small crack in the lower-left region is not correctly localized. In future work, we will further improve the model’s ability to simultaneously segment multiple anomalous regions.

\section{Detailed ZSAS results}
\label{secD}
In Tables \ref{AP} to \ref{IoU_nor}, we present the detailed quantitative results for each specific category of the MVTec-AD datasets. In Figures \ref{fig_r7} to \ref{fig_r19}, we provide more extensive qualitative results for various categories across different industrial and medical datasets. Moreover, we provide additional examples of the three instruction types: Direct Segmentation, Describe-then-Segment, and Segment-then-Explain, in Figures \ref{fig_r3} through \ref{fig_r6}.

\section{Limitations}
\label{secE}
Although AG-VAS achieves state-of-the-art ZSAS performance across multiple medical and industrial datasets and produces highly accurate binary segmentation masks, it still has two main limitations. First, despite its ability to leverage richer contextual semantics and world knowledge during inference to enhance zero-shot generalization, LMM-based methods inherently suffer from slower inference speed, which constrains their applicability in real-time industrial environments. Second, while the segment-then-explain and describe-then-segment modes improve human–model interaction and provide greater interpretability, they do not yield additional performance gains without more expert priors. The extra textual generation tends to divert the model’s focus from the segmentation objective, which is why AG-VAS primarily adopts implicit reasoning mode to directly output segmentation results.

\begin{figure}[h]
	\centering
	\includegraphics[width=1.0\columnwidth]{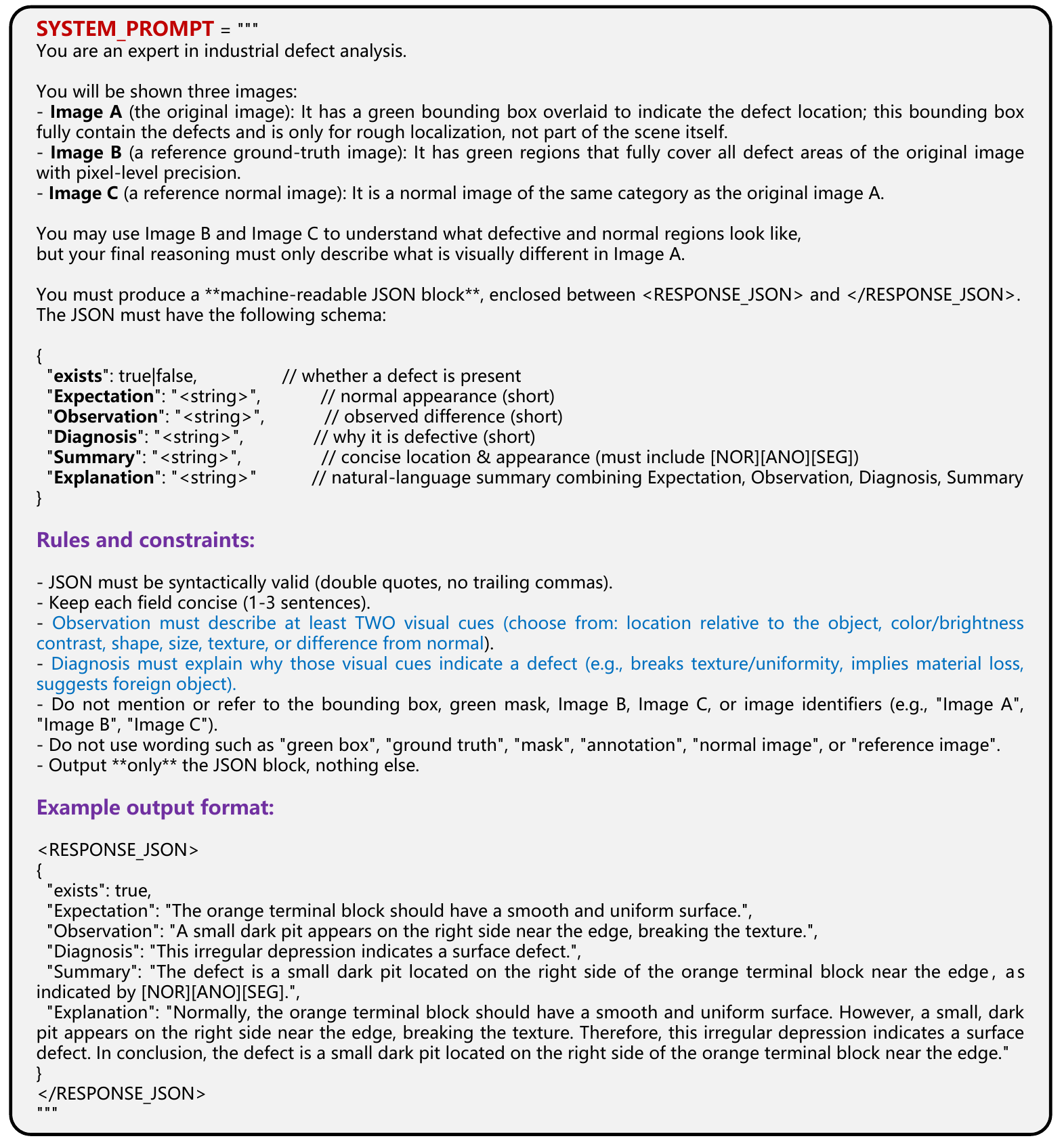}
	\caption{The design of system prompts.}
	\label{system_prompts}
\end{figure}

\begin{figure}[h]
	\centering
	\includegraphics[width=0.8\columnwidth]{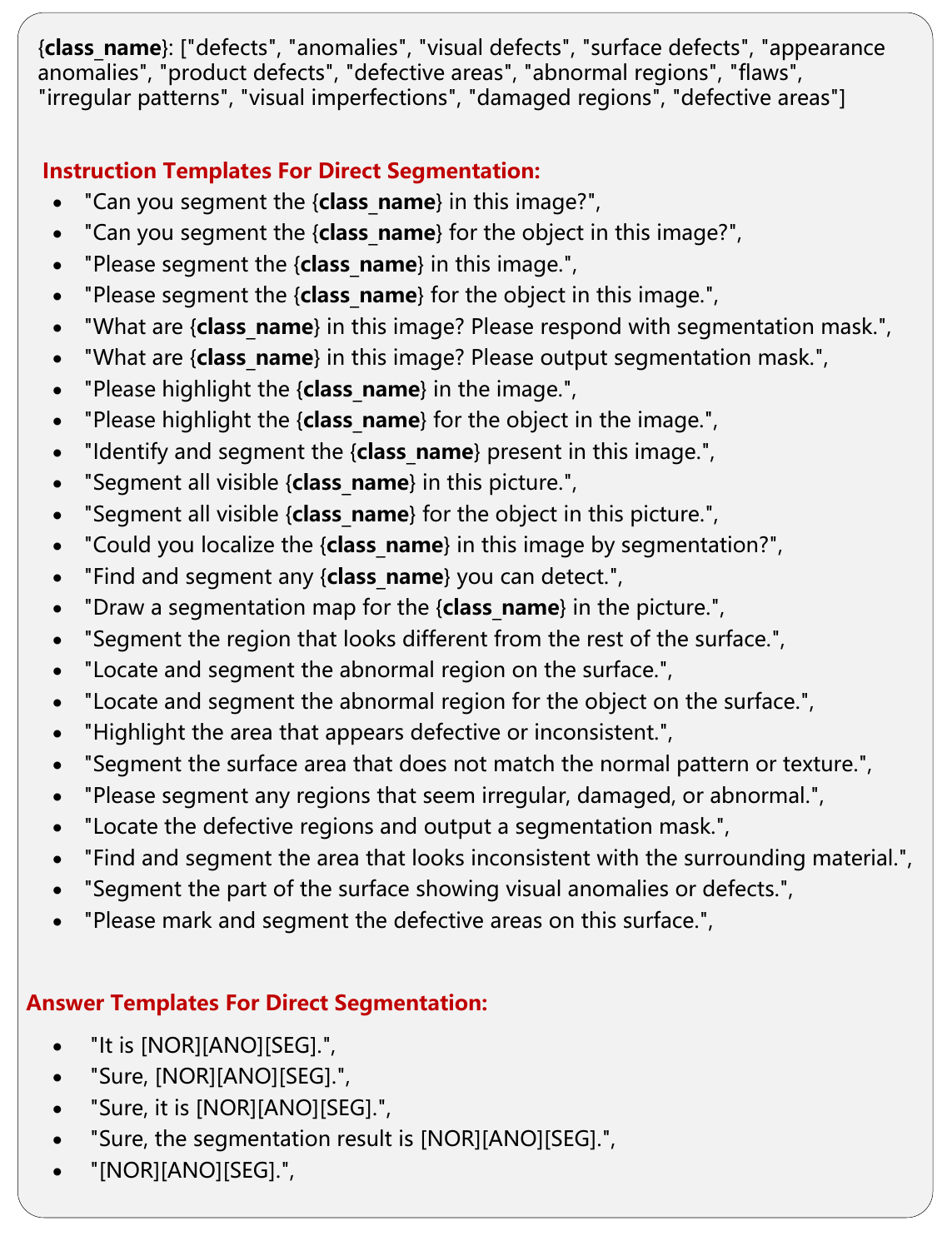}
	\caption{The design of instruction and answer templates for direct segmentation.}
	\label{templates1}
\end{figure}

\begin{figure}[h]
	\centering
	\includegraphics[width=0.8\columnwidth]{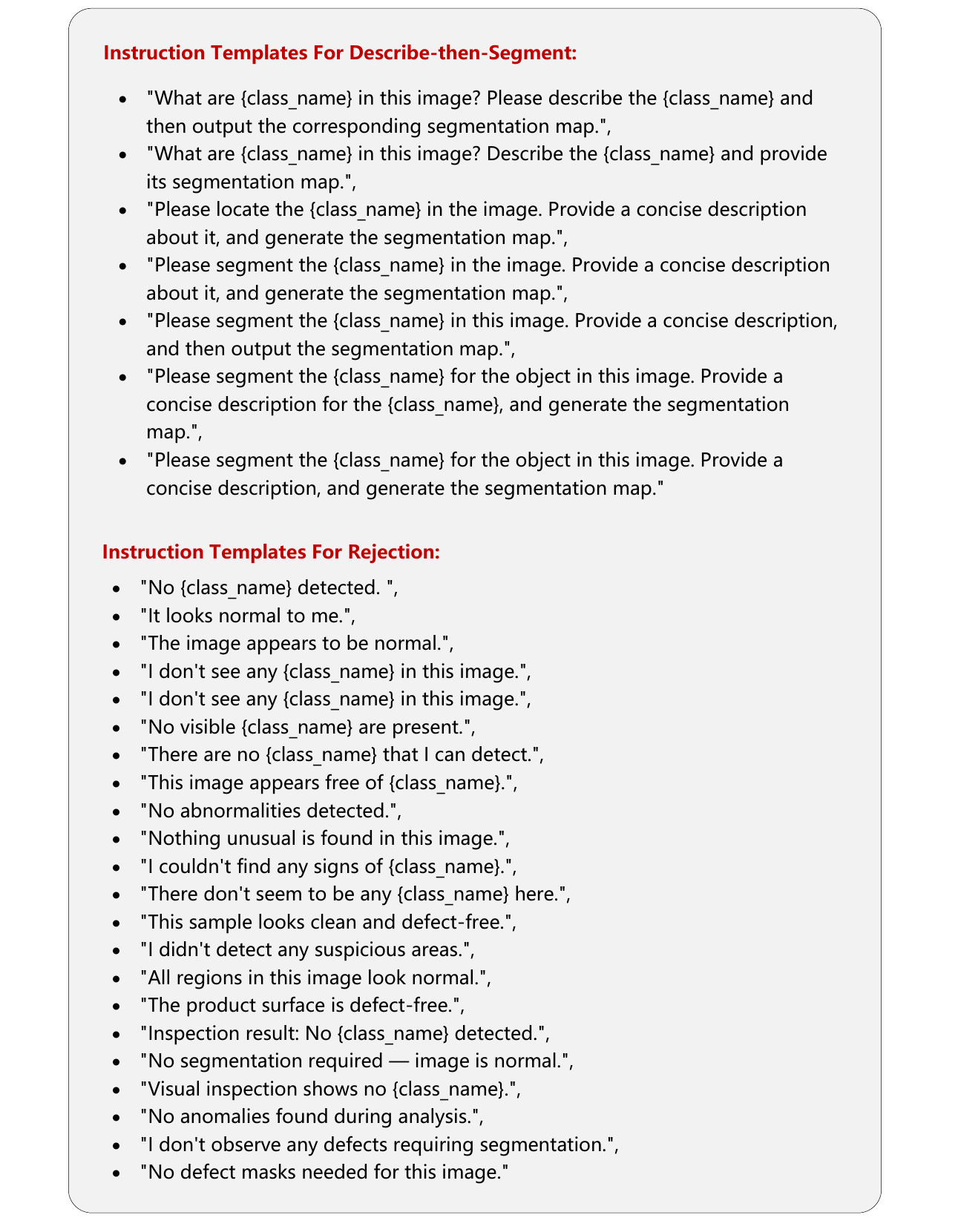}
	\caption{The design of instruction templates for Describe-then-Segment and Rejection.}
	\label{templates2}
\end{figure}

\clearpage
\begin{table*}[t]
	\caption{Comparison of different categories in terms of pixel-level AP on the MVTec-AD.} 
	\centering
	\label{AP}
	\renewcommand{\arraystretch}{1.0}
	\resizebox{1.0\columnwidth}{!}
	{
		\begin{tabular}{ccccccccccc}
			\toprule
			object     & AnomalyCLIP & Bayes-PFL & PaDT & PixelLM & LISA-7B & LISA-13B & LISA-13B* & LISA-OV & LISA-OV* & AG-VAS \\  \midrule
			bottle     & \textbf{55.3}        & 64.7      & 10.1 & 22.9    & 14.2    & 15.6     & 34.3      & 35.0    & 47.5     & 45.8   \\
			cable      & 12.3        & 15.1      & 6.1  & 9.7     & 8.8     & 22.6     & 10.8      & 25.3    & \textbf{27.8}     & 27.1   \\
			capsule    & 27.7        & 34.4      & 4.0  & 7.0     & 15.1    & 15.7     & 14.4      & 14.9    & 17.1     & \textbf{38.4}   \\
			carpet     & 56.6        &\textbf{ 82.2}      & 1.4  & 40.8    & 1.1     & 11.3     & 59.4      & 30.4    & 54.5     & 72.8   \\
			grid       & 24.1        & 41.1      & 0.7  & 0.7     & 0.4     & 0.5      & 60.4      & 23.4    & 57.5     & \textbf{64.1}   \\
			hazelnut   & 43.4        & 70.9      & 5.8  & 7.6     & 11.5    & 11.7     & 36.0      & 34.8    & 55.9     & \textbf{70.2}   \\
			leather    & 22.8        & 58.9      & 0.8  & 0.5     & 3.1     & 2.5      &\textbf{ 77.0 }     & 58.6    & 69.9     & 76.7   \\
			metal\_nut & \textbf{26.5}        & 24.3      & 25.0 & 14.5    & 19.9    & 21.2     & 19.9      & \textbf{38.5}    & 19.3     & 23.3   \\
			pill       & 34.1        & 29.4      & 9.6  & 31.6    & 15.6    & \textbf{40.8 }    & 20.6      & 32.0    & 25.3     & 25.5   \\
			screw      & 27.5        & \textbf{41.6}      & 2.3  & 2.9     & 4.9     & 4.0      & 4.2       & 13.0    & 12.2     & 33.7   \\
			tile       & 61.7        & 75.9      & 8.5  & 8.1     & 3.7     & 4.9      & 85.3      & 69.1    & 79.8     & \textbf{89.6}   \\
			toothbrush & 19.4        & 34.0      & 4.8  & 21.7    & 5.9     & 5.8      & 28.0      & 9.3     & 25.0     & \textbf{53.4}   \\
			transistor & 15.6        & 12.9      & 9.2  & 16.9    & 6.7     & 7.4      & 11.1      & 12.7    & 15.6     & \textbf{19.6}   \\
			wood       & 52.7        & 70.5      & 4.5  & 15.5    & 31.5    & 42.6     &\textbf{ 80.8}      & 69.2    & 84.1     & 74.1   \\
			zipper     & 38.7        & 69.0      & 8.5  & 3.1     & 4.5     & 11.9     & 21.8      & 20.0    & 22.9     & \textbf{51.2}   \\   \midrule
			mean       & 34.5        & 48.3      & 6.8  & 13.6    & 9.8     & 14.6     & 37.6      & 32.4    & 41.0     & \textbf{51.0   }\\ \bottomrule
		\end{tabular}
	}
\end{table*}

\begin{table*}[t]
	\caption{Comparison of different categories in terms of pixel-level F1-Max on the MVTec-AD.} 
	\centering
	\label{F1-Max}
	\renewcommand{\arraystretch}{1.0}
	\resizebox{1.0\columnwidth}{!}
	{
		\begin{tabular}{ccccccccccc}
			\toprule
			object     & AnomalyCLIP & Bayes-PFL & PaDT & PixelLM & LISA-7B & LISA-13B & LISA-13B* & LISA-OV & LISA-OV* & AG-VAS \\   \midrule
			bottle     & 51.6        &\textbf{ 61.3}      & 18.9 & 34.6    & 27.4    & 30.5     & 38.3      & 38.3    & 46.4     & 47.0   \\
			cable      & 18.9        & 23.5      & 14.8 & 16.4    & 13.8    & 27.3     & 17.1      & 36.9    & 38.0     & \textbf{34.5}   \\
			capsule    & 31.0        & 40.5      & 7.9  & 13.2    & 27.2    & 26.3     & 27.8      & 26.7    & 27.8     & \textbf{46.5}   \\
			carpet     & 57.1        & 74.9      & 3.7  & 43.7    & 3.2     & 18.6     & 57.2      & 37.7    & 52.4     & \textbf{66.3  } \\
			grid       & 32.0        & 41.6      & 1.6  & 1.5     & 1.4     & 1.4      & 61.7      & 29.1    & 59.2     & \textbf{64.2 }  \\
			hazelnut   & 47.6        &\textbf{ 66.2 }     & 11.3 & 15.5    & 23.0    & 23.2     & 44.3      & 36.3    & 56.9     & 65.0   \\
			leather    & 33.2        & 56.9      & 1.9  & 1.4     & 4.9     & 5.3      & \textbf{70.4}      & 62.1    & 64.4     & 69.2   \\
			metal\_nut & 33.1        & 35.5      & 36.2 & 36.3    & 37.9    & 39.1     & 21.0      & \textbf{53.4}    & 21.0     & 22.6   \\
			pill       & 35.5        & 32.9      & 17.4 & 28.7    & 24.1    & 43.8     & 27.0      & 33.2    & 31.4     & 33.9   \\
			screw      & 33.4        & 44.1      & 4.8  & 5.9     & 10.5    & 8.7      & 9.6       & 23.2    & 26.9     & \textbf{44.7}   \\
			tile       & 64.9        & 71.2      & 18.4 & 15.4    & 13.2    & 13.5     & 82.6      & 68.2    & 77.5     & \textbf{85.3}   \\
			toothbrush & 29.0        & 38.0      & 10.5 & 27.4    & 11.9    & 11.7     & 34.7      & 15.6    & 31.8     & \textbf{58.7}   \\
			transistor & 18.8        & 17.3      & 20.0 & 21.9    & 9.3     & 14.2     & 18.3      & 16.9    & 21.6     & \textbf{28.7}   \\
			wood       & 55.2        & 65.5      & 10.6 & 26.4    & 34.8    & 41.3     & 74.8      & 67.8    & 77.2     & \textbf{72.4}   \\
			zipper     & 45.0        & \textbf{66.9}      & 15.9 & 6.3     & 10.3    & 22.4     & 27.8      & 29.6    & 28.4     & 52.2   \\  \midrule
			mean       & 39.1        & 49.1      & 12.9 & 19.6    & 16.9    & 21.8     & 40.9      & 38.3    & 44.1     & \textbf{52.7}     \\ \bottomrule
		\end{tabular}
	}
\end{table*}

\begin{table*}[t]
	\caption{Comparison of different categories in terms of $\text{IoU}_{\text{ano}}$ on the MVTec-AD.} 
	\centering
	\label{IoU_ano}
	\renewcommand{\arraystretch}{1.0}
	\resizebox{1.0\columnwidth}{!}
	{
		\begin{tabular}{ccccccccccc}
			\toprule
			object     & AnomalyCLIP & Bayes-PFL & PaDT & PixelLM & LISA-7B & LISA-13B & LISA-13B* & LISA-OV & LISA-OV* & AG-VAS \\  \midrule
			bottle     & 27.3        & 25.2      & 20.0 & 26.7    & 20.0    & 25.4     & 30.7      & 36.7    & 23.8     & \textbf{41.4  } \\
			cable      & 10.3        & 7.5       & 11.5 & 5.0     & 3.4     & 8.9      & 4.0       & 23.4    & 16.0     & \textbf{29.2 }  \\
			capsule    & 16.8        & 7.6       & 4.7  & 4.9     & 5.4     & 8.3      & 16.0      & 14.1    & 12.1     & \textbf{36.3}   \\
			\textbf{	carpet}     & 32.5        & 48.9      & 16.4 & 29.6    & 3.3     & 13.6     & 46.5      & 39.0    & 43.6     & 52.1   \\
			grid       & 20.5        & 24.4      & 12.9 & 2.0     & 0.8     & 0.5      & \textbf{43.7}      & 32.8    & 47.9     & 41.7   \\
			hazelnut   & 28.0        & 9.4       & 14.5 & 13.2    & 17.6    & 23.0     & 45.8      & 44.7    & 45.7     & \textbf{64.4 }  \\
			leather    & 13.3        & 16.3      & 26.8 & 1.5     & 5.3     & 1.7      & \textbf{56.0}      & 54.0    & 58.4     & 48.0   \\
			metal\_nut & 21.7        & 24.0      & 26.2 & 23.0    & 22.7    & 13.1     & 38.5      & 39.8    & 34.6     & \textbf{46.8}   \\
			pill       & 20.5        & 9.2       & 10.8 & 10.5    & 9.0     & 19.4     & 18.7      & 29.4    & 31.2     & \textbf{54.5}   \\
			screw      & 13.9        & 1.9       & 2.9  & 2.9     & 2.8     & 4.5      & 6.0       & 12.2    & 9.3      & \textbf{31.9}   \\
			tile       & 43.2        & 51.7      & 46.9 & 19.1    & 5.3     & 14.9     & 74.1      & 58.7    & 61.6     & \textbf{75.0 }  \\
			toothbrush & 20.4        & 6.9       & 7.2  & 4.8     & 7.3     & 6.6      & 13.1      & 7.7     & 24.3     & \textbf{43.7 }  \\
			transistor & 14.7        & 13.8      & 9.9  & 8.6     & 5.5     & 0.5      & 7.0       & 12.9    & 13.0     & \textbf{18.8}   \\
			wood       & 33.8        & 39.5      & 22.7 & 23.7    & 41.1    & 51.9     & 63.9      & 59.7    & 50.9     & \textbf{62.3}   \\
			zipper     & 22.1        & \textbf{46.9}      & 13.0 & 3.6     & 2.7     & 13.7     & 12.7      & 18.5    & 11.7     & 26.0   \\ \midrule
			mean       & 22.6        & 22.2      & 16.4 & 11.9    & 10.2    & 13.7     & 31.8      & 32.3    & 32.3     & \textbf{44.8 }   \\ \bottomrule
		\end{tabular}
	}
\end{table*}

\begin{table*}[t]
	\caption{Comparison of different categories in terms of $\text{IoU}_{\text{nor}}$ on the MVTec-AD.} 
	\centering
	\label{IoU_nor}
	\renewcommand{\arraystretch}{1.0}
	\resizebox{1.0\columnwidth}{!}
	{
		\begin{tabular}{ccccccccccc}
			\toprule
			object     & AnomalyCLIP & Bayes-PFL & PaDT & PixelLM & LISA-7B & LISA-13B & LISA-13B* & LISA-OV & LISA-OV* & AG-VAS \\  \midrule
			bottle     & 0.0         & 0.0       & 0.0  & 0.0     & 0.0     & 0.0      & 10.0      & 0.0     & 75.0     & \textbf{85.0}   \\
			cable      & 17.2        & 0.0       & 0.0  & 34.5    & 25.9    & 17.2     & 0.0       & 0.0     & 56.9     & \textbf{55.2}   \\
			capsule    & 73.9        & 0.0       & 0.0  & 0.0     & 0.0     & 0.0      & 0.0       & 0.0     & 91.3     & \textbf{95.7}   \\
			carpet     & 78.6        & 96.4      & 0.0  & 0.0     & 0.0     & 3.6      & \textbf{100.0}     & 57.1    & 89.3     & \textbf{100.0 } \\
			grid       & 76.2        & 95.2      & 0.0  & 0.0     & 0.0     & 0.0      & \textbf{95.2}      & 0.0     & \textbf{95.2}     & \textbf{95.2}   \\
			hazelnut   & 7.5         & 0.0       & 0.0  & 0.0     & 0.0     & 0.0      & 0.0       & 0.0     & 65.0     & \textbf{97.5  } \\
			leather    & 28.1        & 53.1      & 0.0  & 0.0     & 0.0     & 0.0      & 90.6      & 0.0     & \textbf{96.9  }   & \textbf{96.9}   \\
			metal\_nut & 9.1         & 0.0       & 0.0  & 0.0     & 0.0     & 0.0      & 36.4      & 0.0     & 95.5     & \textbf{95.5}   \\
			pill       & 11.5        & 0.0       & 0.0  & 0.0     & 0.0     & 0.0      & \textbf{100.0}     & 0.0     & 53.9     & 96.2   \\
			screw      & 48.8        & 0.0       & 0.0  & 0.0     & 2.4     & 2.4      & 4.9       & 0.0     & 90.2     & 53.7   \\
			tile       & 78.8        & 12.1      & 0.0  & 0.0     & 0.0     & 0.0      & 78.8      & 3.0     & 81.8     & \textbf{97.0 }  \\
			toothbrush & 0.0         & 0.0       & 0.0  & 0.0     & 0.0     & 0.0      & 0.0       & 0.0     & 50.0     & \textbf{100.0  }\\
			transistor & 5.0         & 0.0       & 0.0  & 1.7     & 0.0     & 71.7     & 0.0       & 0.0     & 90.0     & \textbf{53.3}   \\
			wood       & 36.8        & 5.3       & 0.0  & 0.0     & 0.0     & 0.0      & 68.4      & 0.0     & 89.5     & \textbf{100.0}  \\
			zipper     & \textbf{100.0}       & 43.8      & 0.0  & 0.0     & 0.0     & 0.0      & 68.8      & 0.0     & 93.8     & 93.8   \\   \midrule
			mean       & 38.1        & 20.4      & 0.0  & 2.4     & 1.9     & 6.3      & 43.5      & 4.0     & 81.0     & \textbf{87.7}    \\ \bottomrule
		\end{tabular}
	}
\end{table*}

\begin{figure*}[h]
	\centering
	\includegraphics[width=0.9\columnwidth]{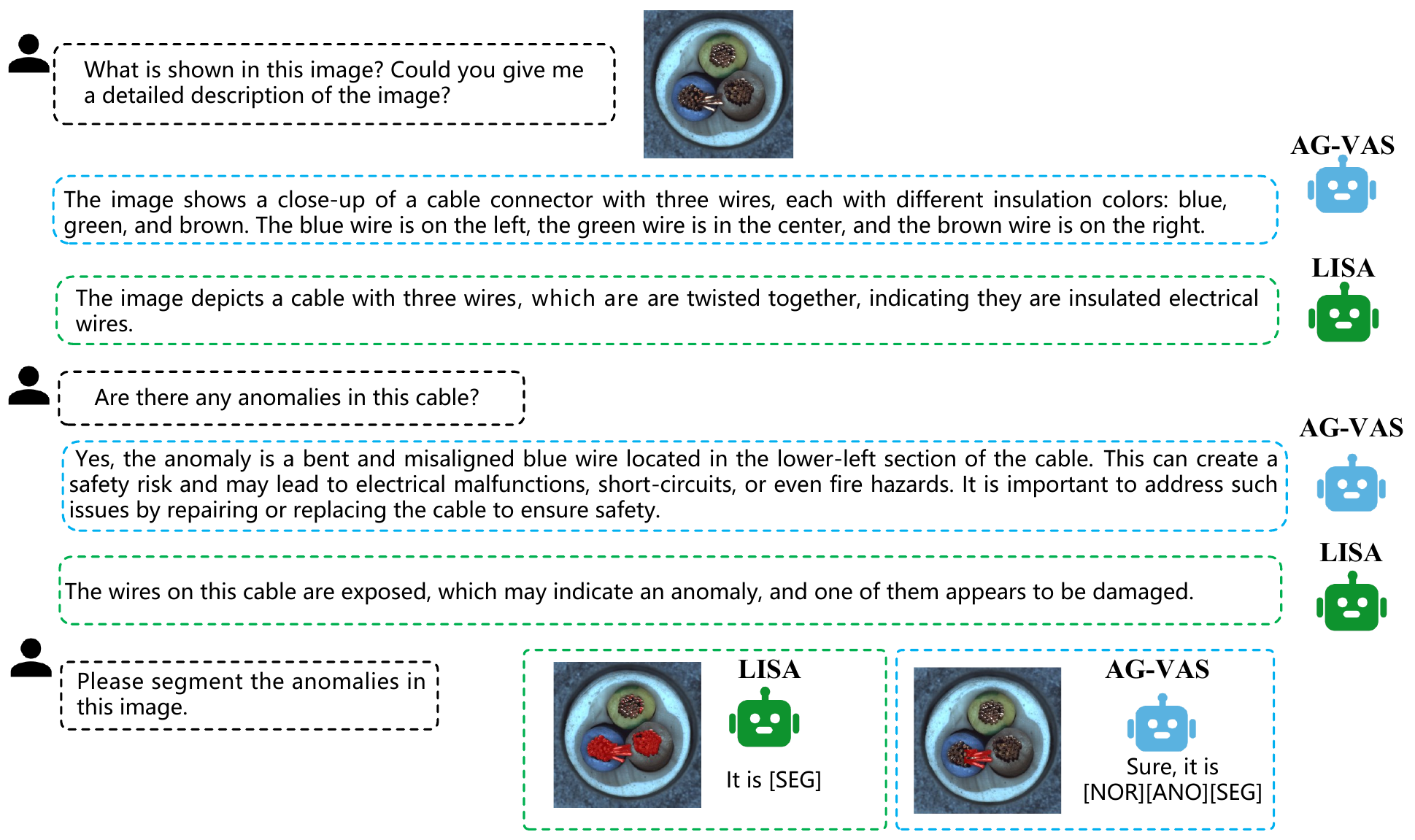}
	\caption{Comparison of the dialogue and segmentation capabilities between the proposed AG-VAS and LISA \cite{LISA}.}
	\label{fig_r1}
\end{figure*}

\begin{figure*}[h]
	\centering
	\includegraphics[width=0.9\columnwidth]{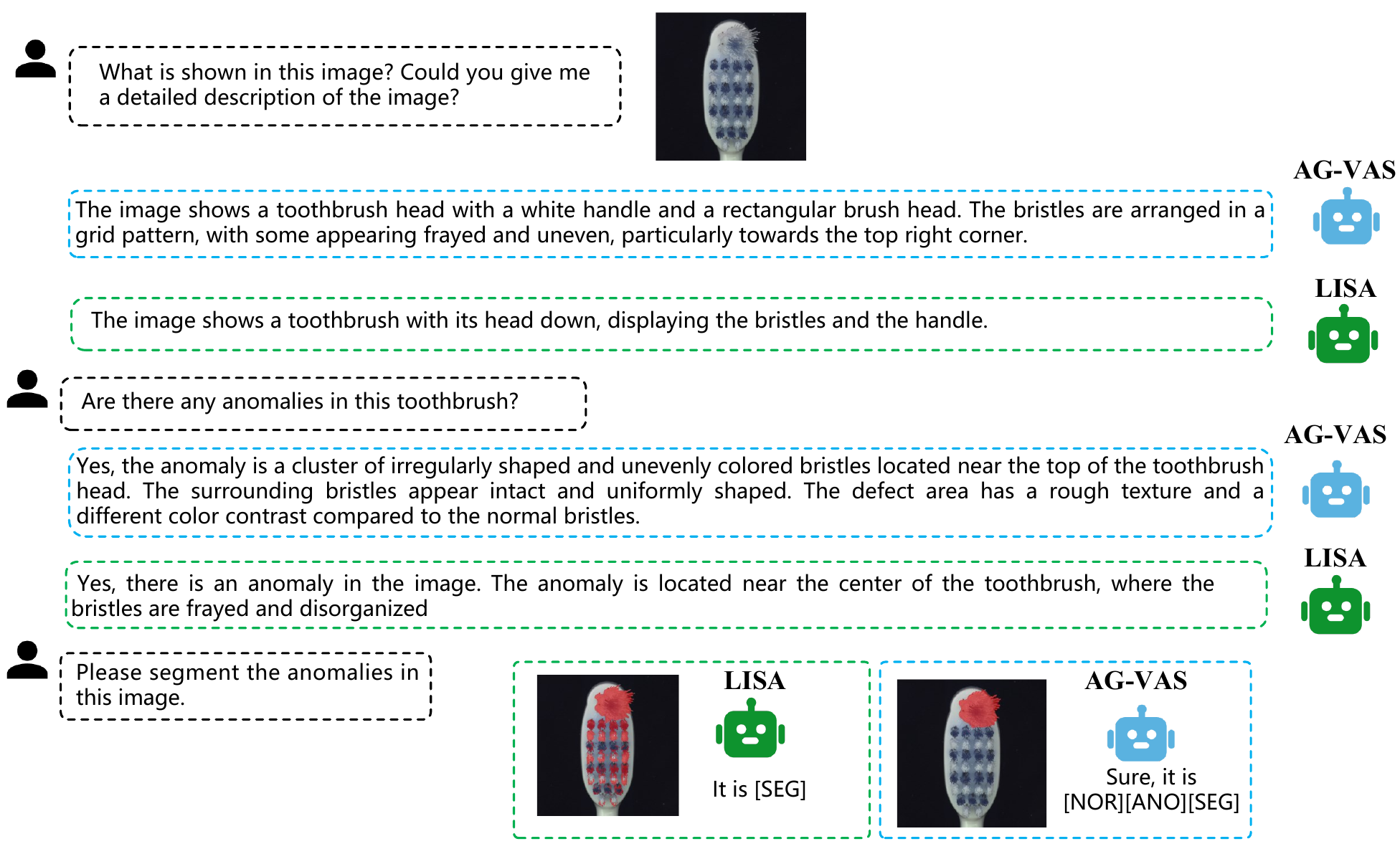}
	\caption{Comparison of the dialogue and segmentation capabilities between the proposed AG-VAS and LISA \cite{LISA}.}
	\label{fig_r2}
\end{figure*}

\begin{figure*}[h]
	\centering
	\includegraphics[width=0.9\columnwidth]{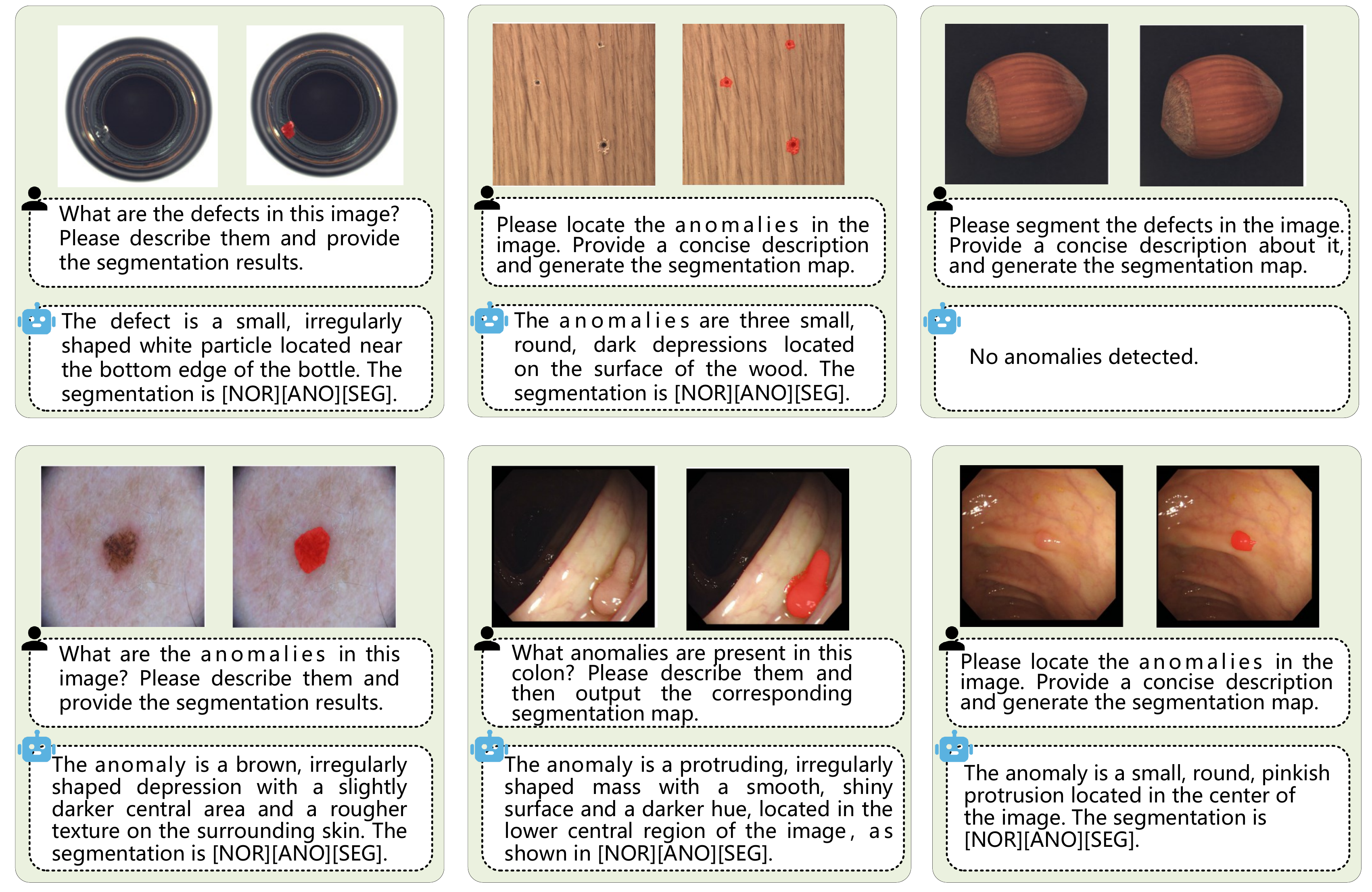}
	\caption{Some examples under the describe-then-segment task. Left: original images; Right: AG-VAS segmentation results.}
	\label{fig_r3}
\end{figure*}

\begin{figure*}[h]
	\centering
	\includegraphics[width=0.9\columnwidth]{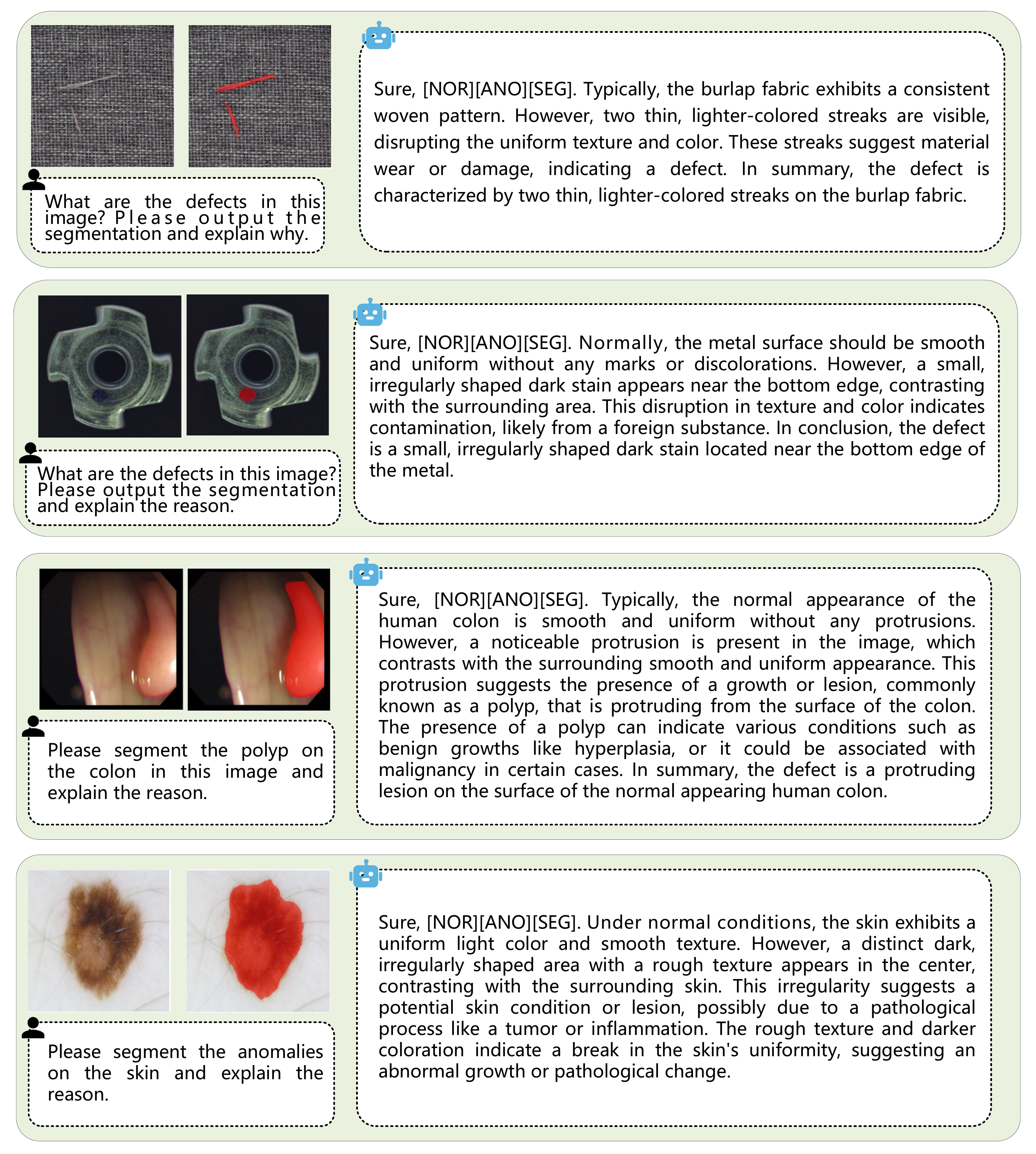}
	\caption{Some examples under the segment-then-explain task. Left: original images; Right: AG-VAS segmentation results.}
	\label{fig_r4}
\end{figure*}

\begin{figure*}[h]
	\centering
	\includegraphics[width=0.9\columnwidth]{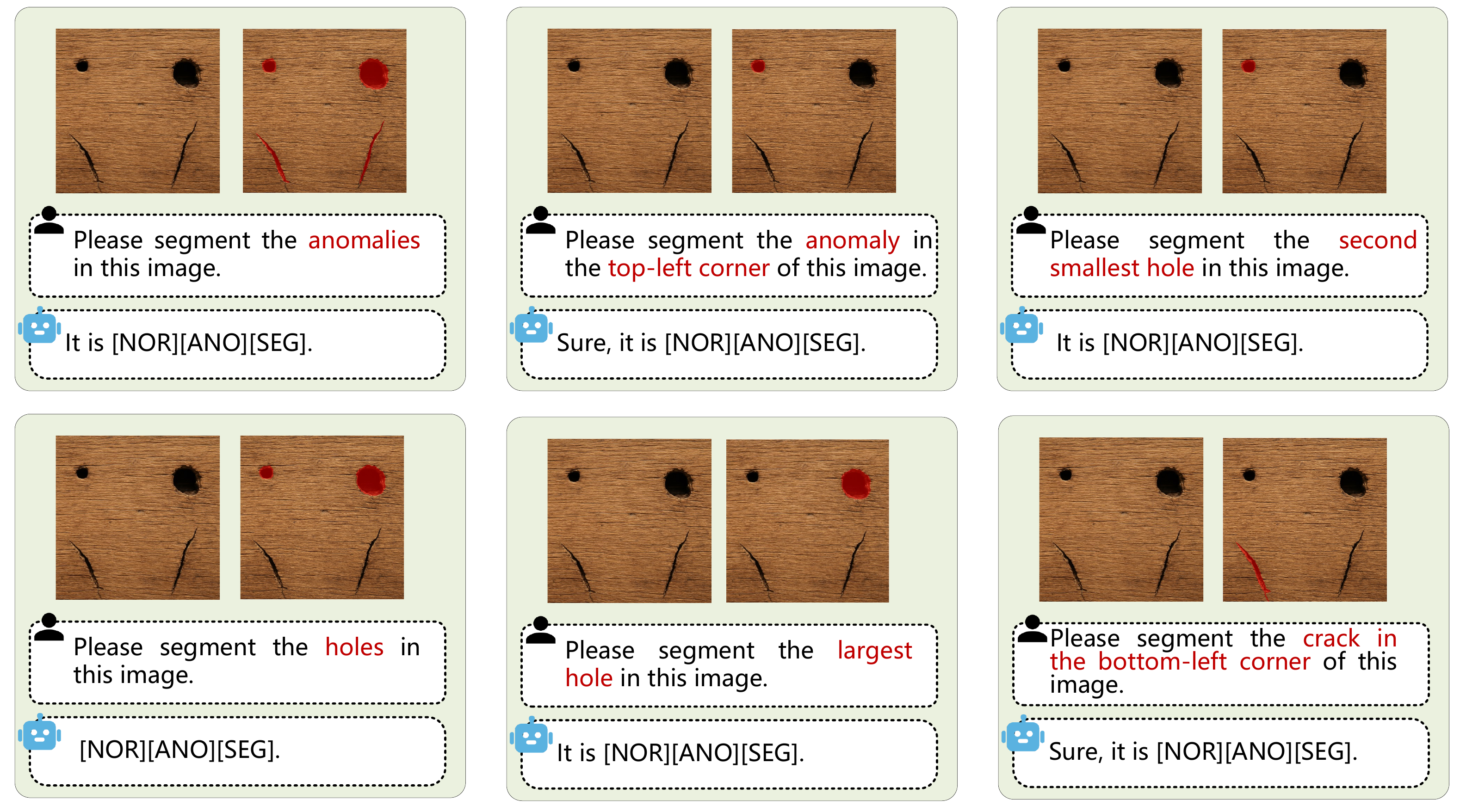}
	\caption{Generalization of AG-VAS to fine-grained segmentation instructions.
		Left: original images; Right: AG-VAS segmentation results.
		AG-VAS effectively captures both positional cues (e.g., “top-left”) and attribute-specific descriptions (e.g., “the smallest”), achieving precise localization of anomalous regions.}
	\label{fig_r5}
\end{figure*}

\begin{figure*}[h]
	\centering
	\includegraphics[width=0.9\columnwidth]{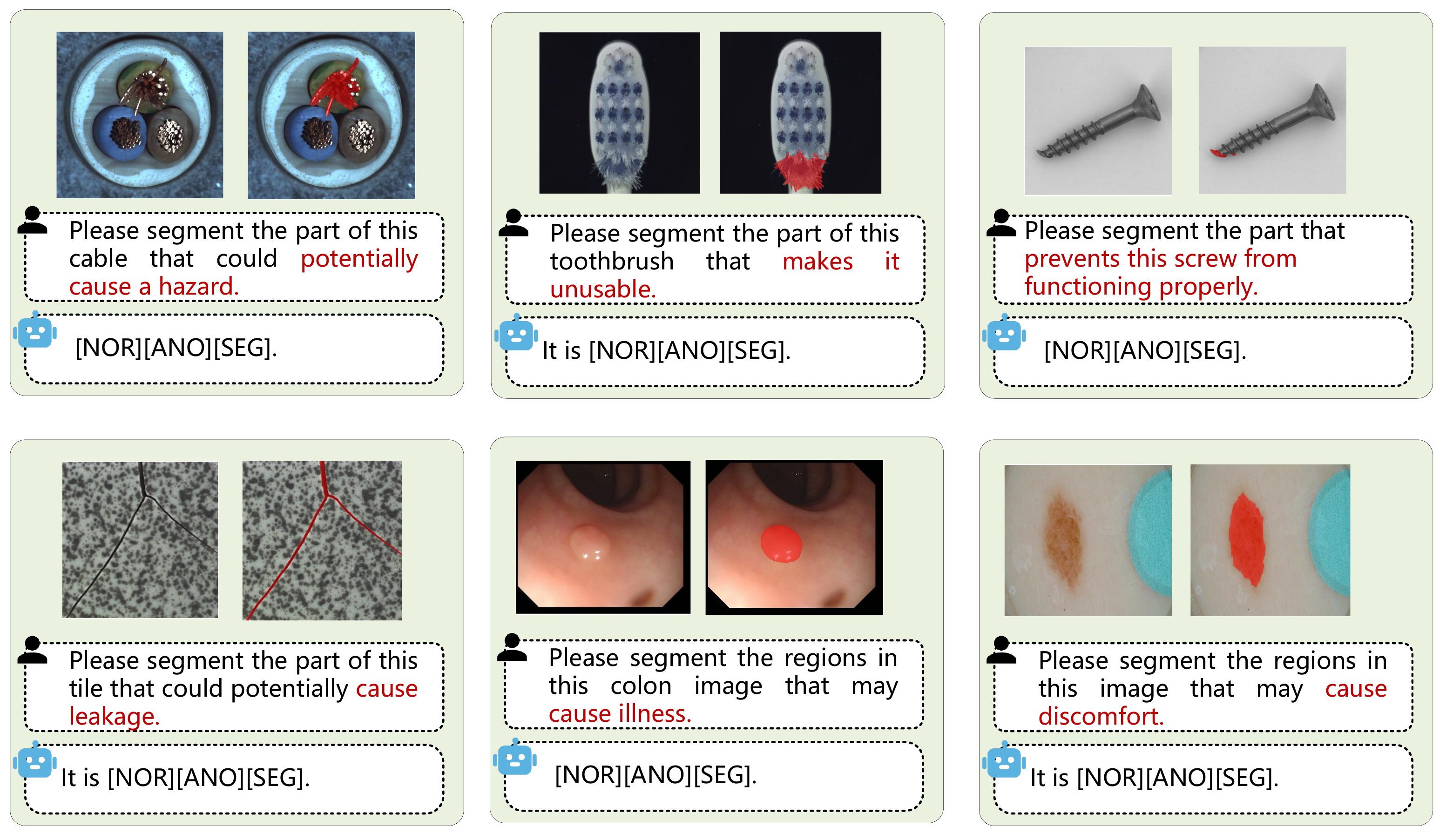}
	\caption{Examples showcasing AG-VAS’s ability to leverage world knowledge for implicit reasoning and direct segmentation.
		Left: original images; Right: AG-VAS segmentation results.
		AG-VAS exploits its learned world knowledge to conduct implicit reasoning and precisely localize anomalous regions.}
	\label{fig_r6}
\end{figure*}

\clearpage
\begin{figure*}[h]
	\centering
	\includegraphics[width=0.9\columnwidth]{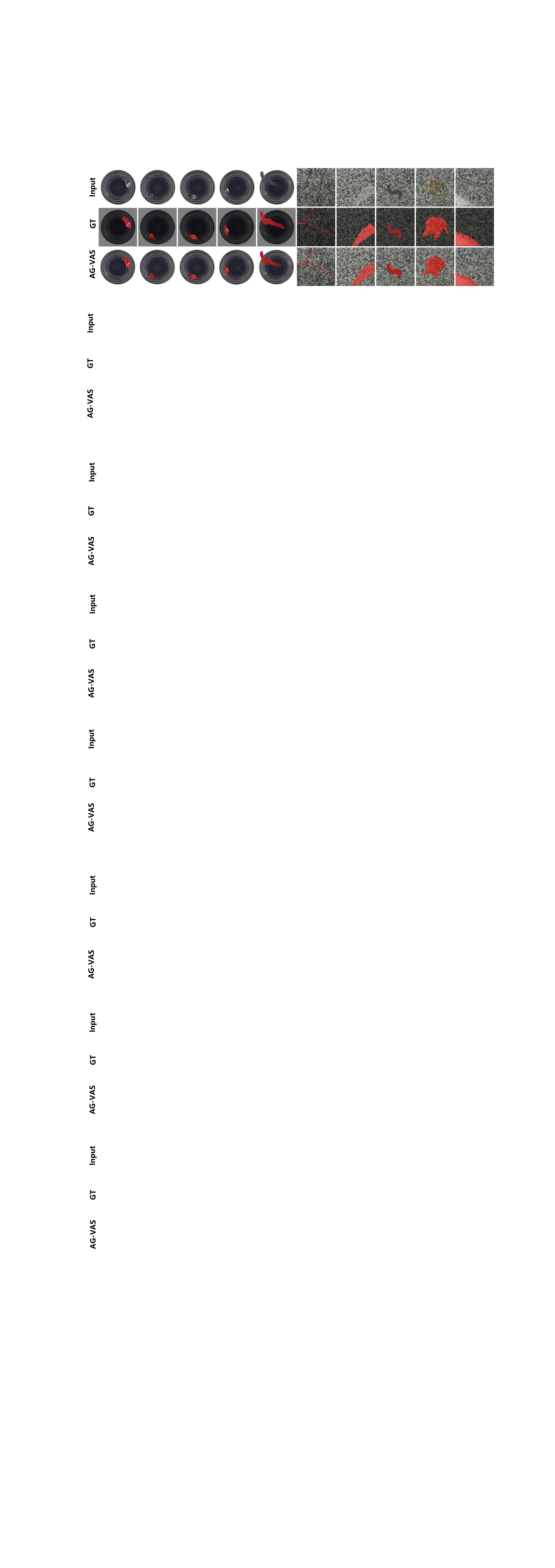}
	\caption{Visualization of segmentation results for the \textbf{bottle} and \textbf{tile} classes on \textbf{MVTecAD}. The top row shows the original input images, the middle row presents the ground-truth masks, and the bottom row displays the binarized segmentation results using the default threshold of 0.5..}
	\label{fig_r7}
\end{figure*}

\begin{figure*}[h]
	\centering
	\includegraphics[width=0.9\columnwidth]{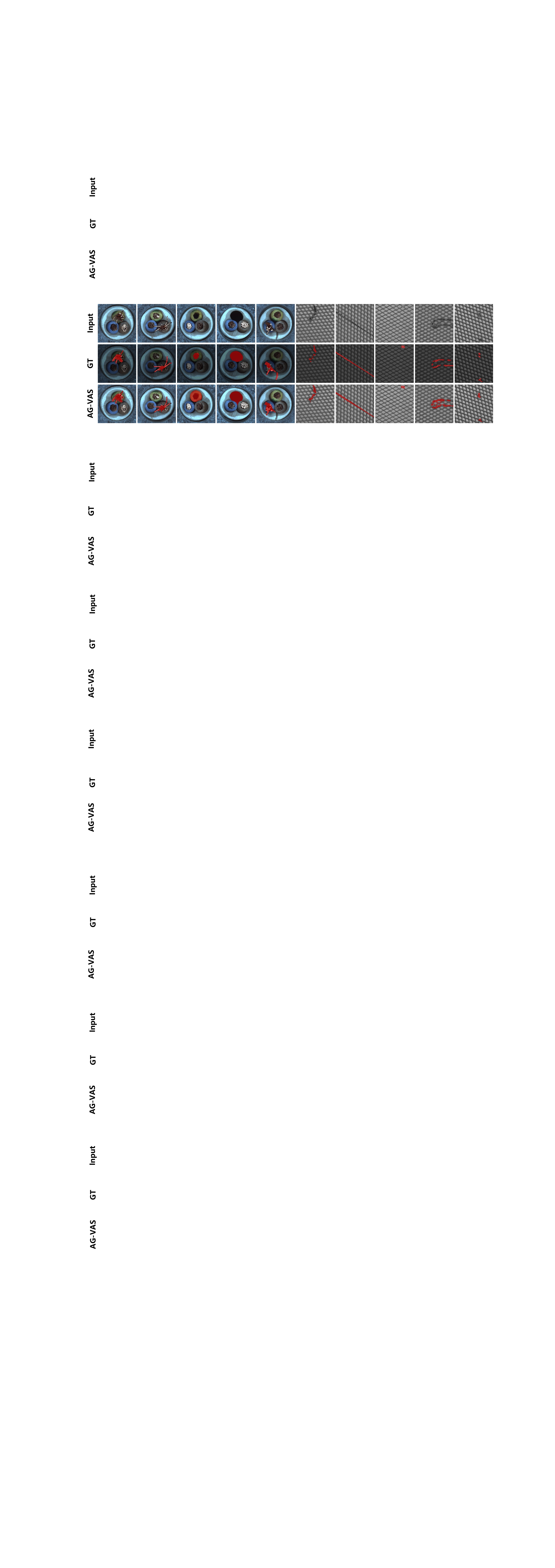}
	\caption{Visualization of segmentation results for the \textbf{cable} and \textbf{grid} classes on \textbf{MVTecAD}. The top row shows the original input images, the middle row presents the ground-truth masks, and the bottom row displays the binarized segmentation results using the default threshold of 0.5.}
	\label{fig_r8}
\end{figure*}

\begin{figure*}[h]
	\centering
	\includegraphics[width=0.9\columnwidth]{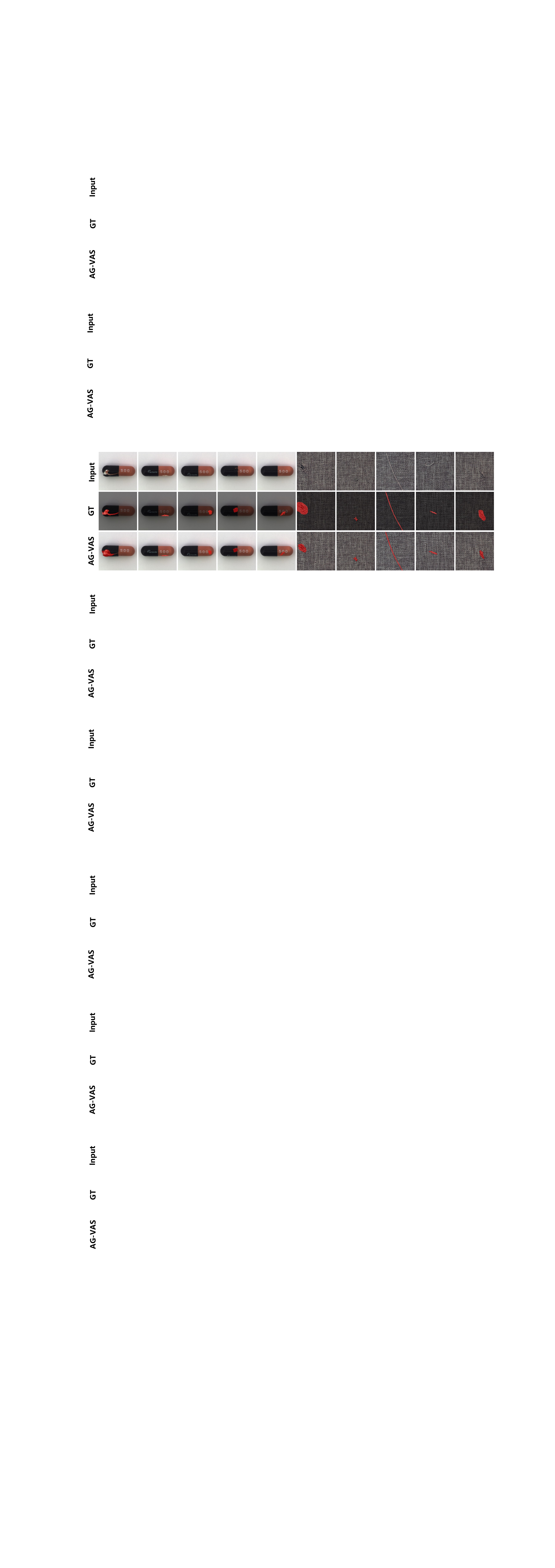}
	\caption{Visualization of segmentation results for the \textbf{capsule} and \textbf{carpet} classes on \textbf{MVTecAD}. The top row shows the original input images, the middle row presents the ground-truth masks, and the bottom row displays the binarized segmentation results using the default threshold of 0.5.}
	\label{fig_r9}
\end{figure*}

\begin{figure*}[h]
	\centering
	\includegraphics[width=0.9\columnwidth]{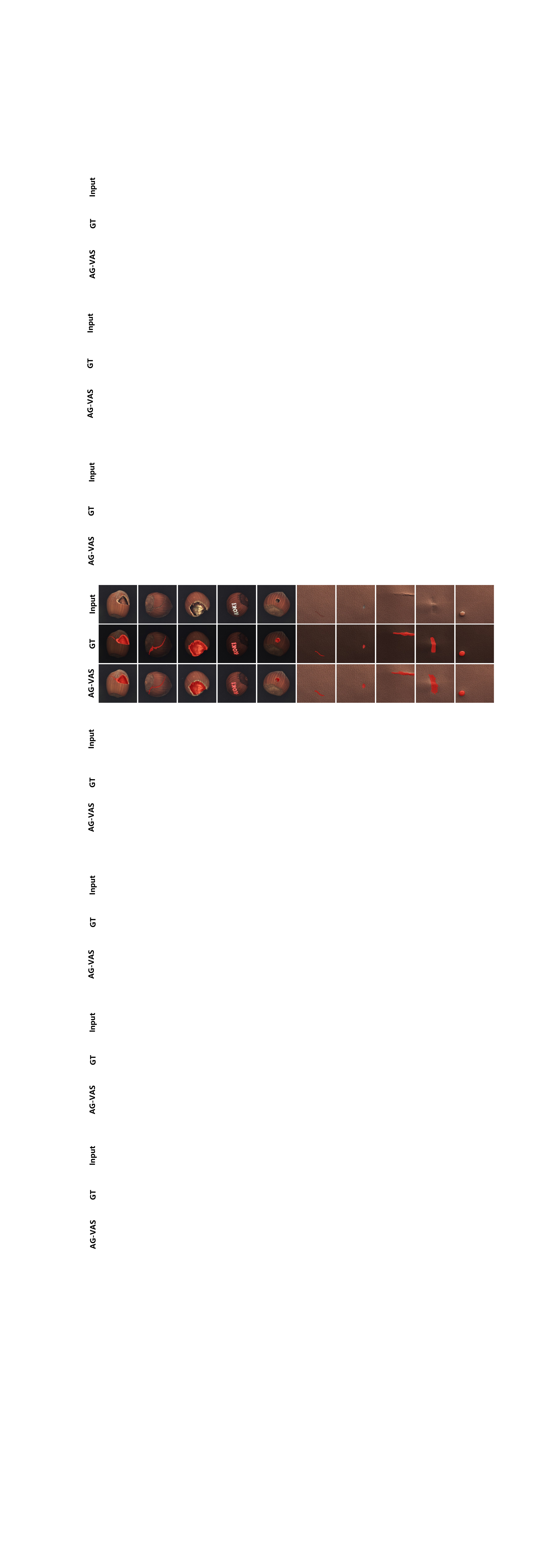}
	\caption{Visualization of segmentation results for the \textbf{hazelnut} and \textbf{leather} classes on \textbf{MVTecAD}. The top row shows the original input images, the middle row presents the ground-truth masks, and the bottom row displays the binarized segmentation results using the default threshold of 0.5.}
	\label{fig_r10}
\end{figure*}

\begin{figure*}[h]
	\centering
	\includegraphics[width=0.9\columnwidth]{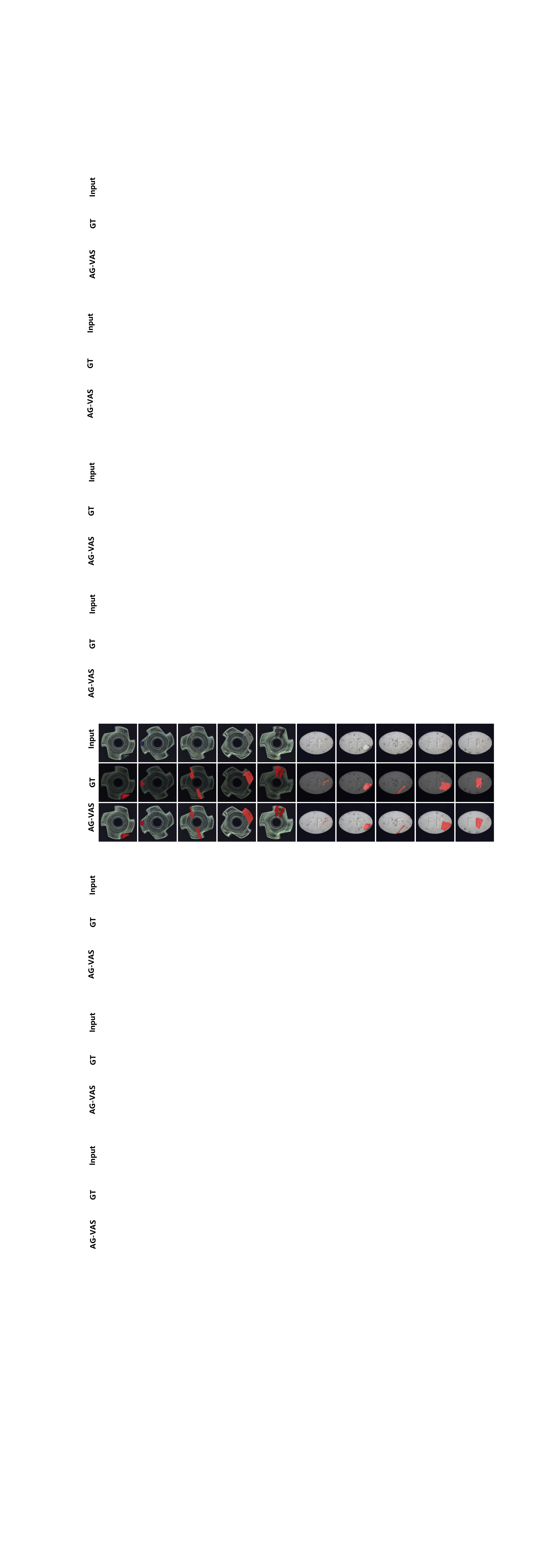}
	\caption{Visualization of segmentation results for the \textbf{metal nut} and \textbf{pill} classes on \textbf{MVTecAD}. The top row shows the original input images, the middle row presents the ground-truth masks, and the bottom row displays the binarized segmentation results using the default threshold of 0.5.}
	\label{fig_r11}
\end{figure*}

\begin{figure*}[h]
	\centering
	\includegraphics[width=0.9\columnwidth]{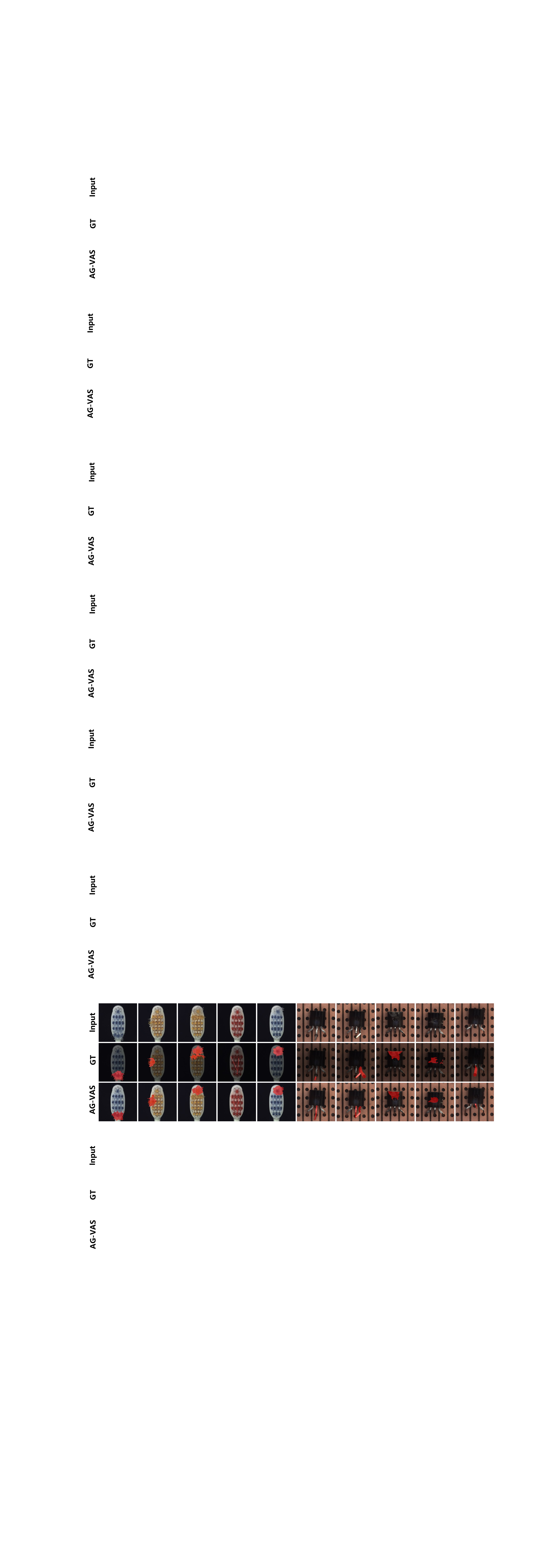}
	\caption{Visualization of segmentation results for the \textbf{toothbrush} and \textbf{transistor} classes on \textbf{MVTecAD}. The top row shows the original input images, the middle row presents the ground-truth masks, and the bottom row displays the binarized segmentation results using the default threshold of 0.5.}
	\label{fig_r12}
\end{figure*}

\begin{figure*}[h]
	\centering
	\includegraphics[width=0.9\columnwidth]{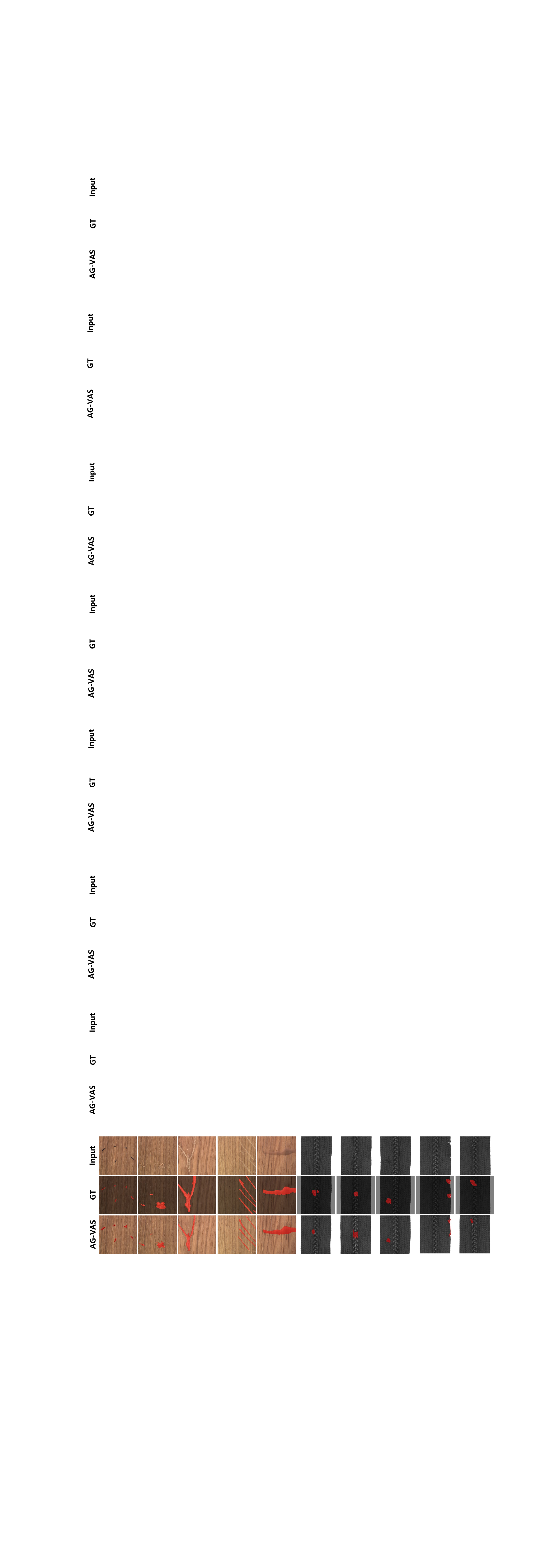}
	\caption{Visualization of segmentation results for the \textbf{wood} and \textbf{ zipper} classes on \textbf{MVTecAD}. The top row shows the original input images, the middle row presents the ground-truth masks, and the bottom row displays the binarized segmentation results using the default threshold of 0.5.}
	\label{fig_r13}
\end{figure*}

\begin{figure*}[h]
	\centering
	\includegraphics[width=0.9\columnwidth]{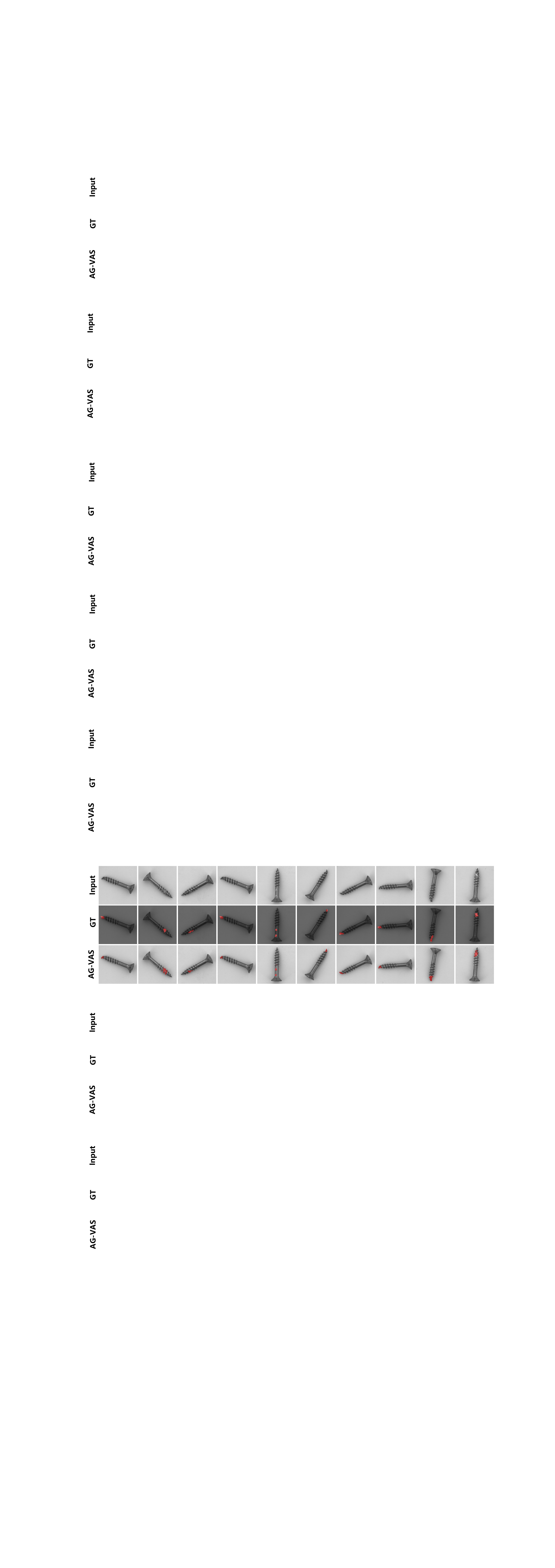}
	\caption{Visualization of segmentation results for the \textbf{screw} class on \textbf{MVTecAD}. The top row shows the original input images, the middle row presents the ground-truth masks, and the bottom row displays the binarized segmentation results using the default threshold of 0.5.}
	\label{fig_r14}
\end{figure*}

\begin{figure*}[h]
	\centering
	\includegraphics[width=0.9\columnwidth]{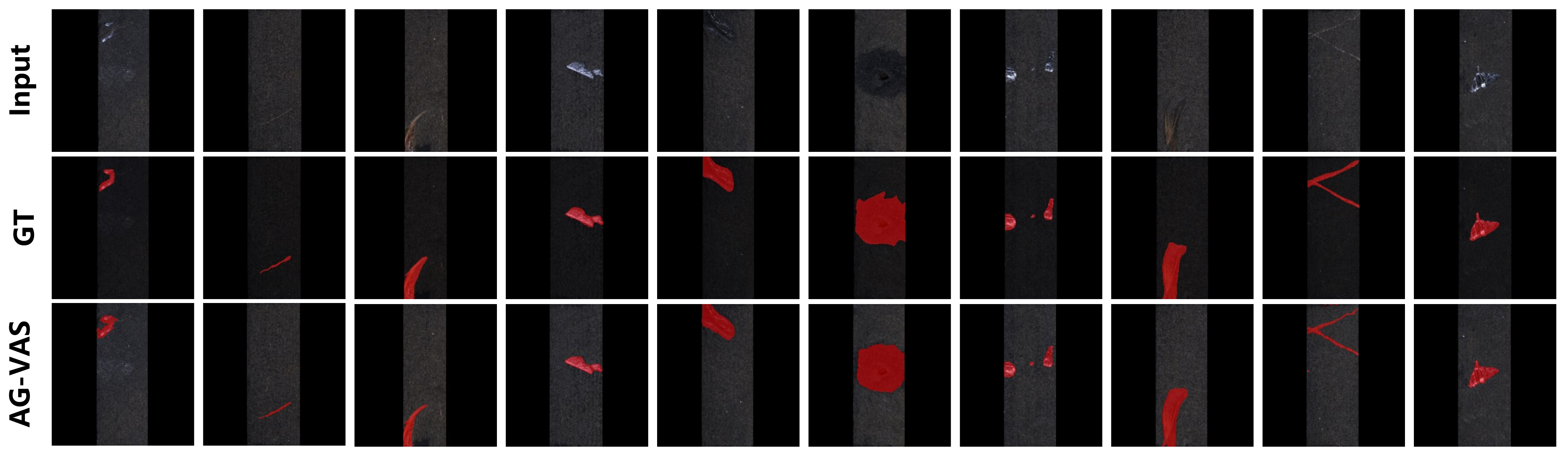}
	\caption{Visualization of segmentation results on \textbf{KSDD2}. The top row shows the original input images, the middle row presents the ground-truth masks, and the bottom row displays the binarized segmentation results using the default threshold of 0.5.}
	\label{fig_r15}
\end{figure*}

\begin{figure*}[h]
	\centering
	\includegraphics[width=0.9\columnwidth]{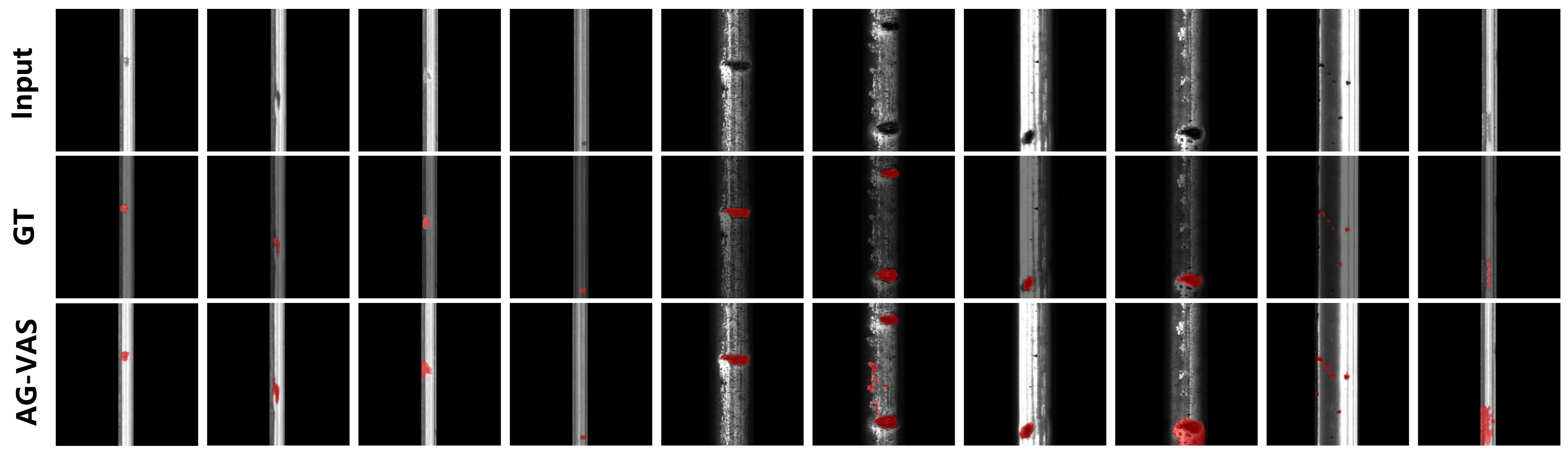}
	\caption{Visualization of segmentation results on \textbf{RSDD}. The top row shows the original input images, the middle row presents the ground-truth masks, and the bottom row displays the binarized segmentation results using the default threshold of 0.5.}
	\label{fig_r16}
\end{figure*}

\begin{figure*}[h]
	\centering
	\includegraphics[width=0.9\columnwidth]{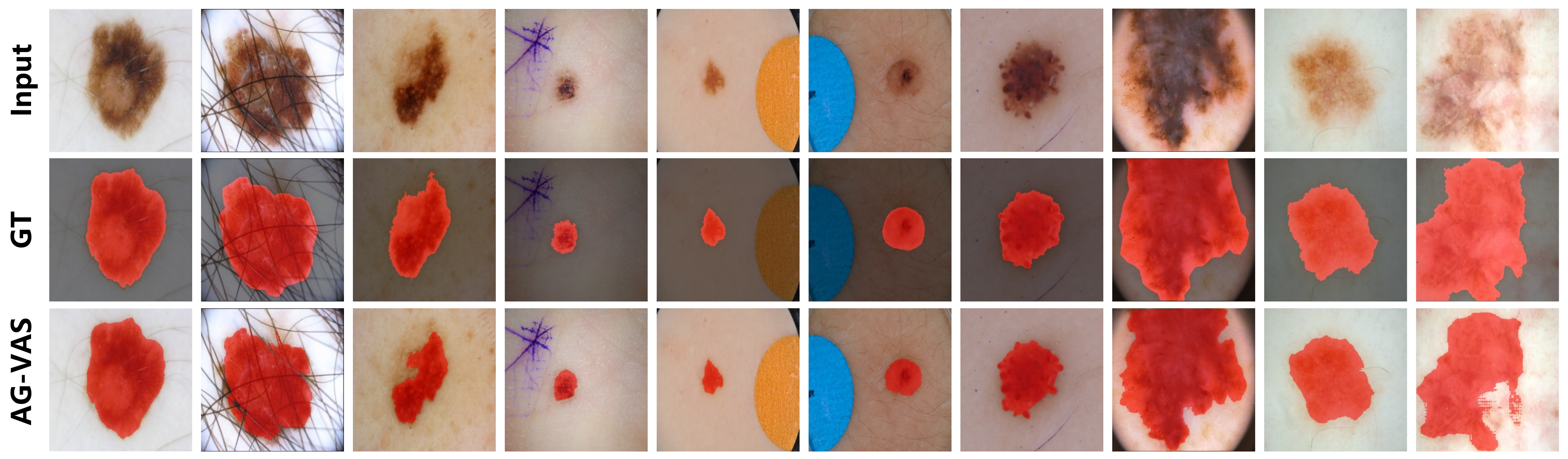}
	\caption{Visualization of segmentation results for the \textbf{skin} class on \textbf{ISIC}. The top row shows the original input images, the middle row presents the ground-truth masks, and the bottom row displays the binarized segmentation results using the default threshold of 0.5.}
	\label{fig_r17}
\end{figure*}

\begin{figure*}[h]
	\centering
	\includegraphics[width=0.9\columnwidth]{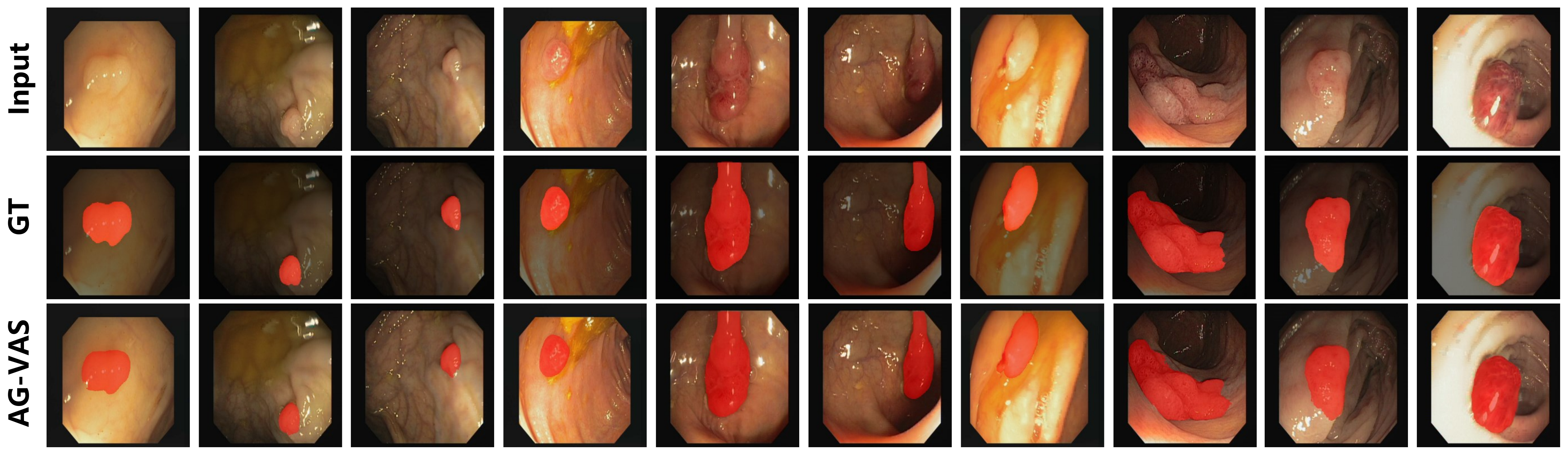}
	\caption{Visualization of segmentation results for the \textbf{colon} class on \textbf{ClinicDB}. The top row shows the original input images, the middle row presents the ground-truth masks, and the bottom row displays the binarized segmentation results using the default threshold of 0.5.}
	\label{fig_r18}
\end{figure*}

\begin{figure*}[h]
	\centering
	\includegraphics[width=0.9\columnwidth]{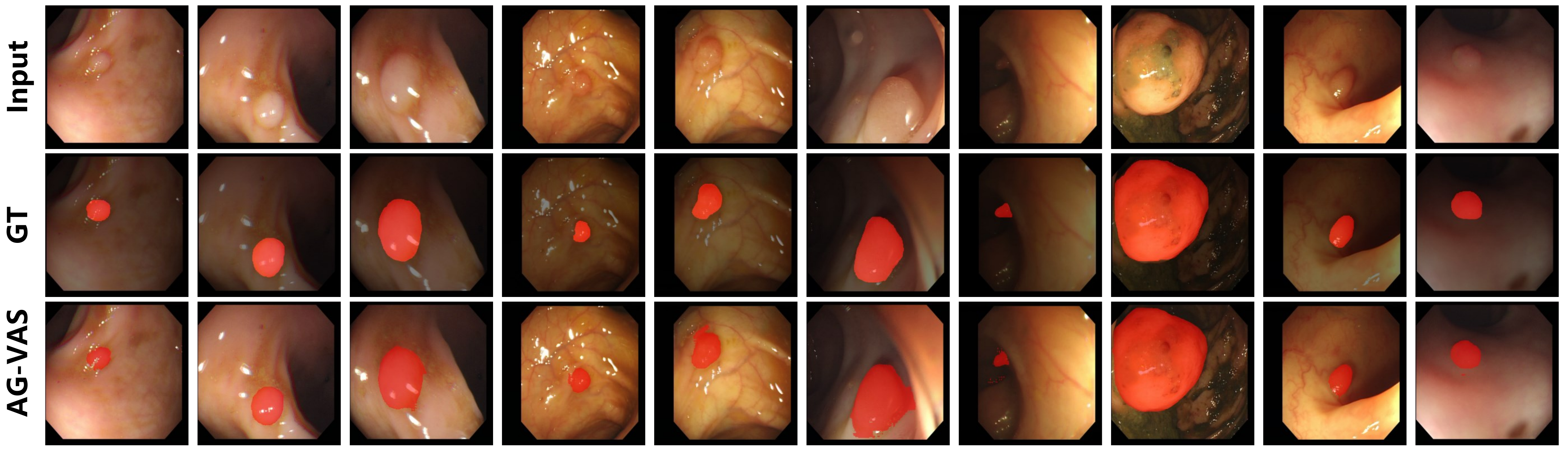}
	\caption{Visualization of segmentation results for the \textbf{colon} class on \textbf{ColonDB}. The top row shows the original input images, the middle row presents the ground-truth masks, and the bottom row displays the binarized segmentation results using the default threshold of 0.5.}
	\label{fig_r19}
\end{figure*}

\end{document}